\documentclass[lettersize,journal]{IEEEtran}
\usepackage{amsmath,amsfonts}
\usepackage{booktabs}
\usepackage{algorithm}
\usepackage[noend]{algpseudocode}
\usepackage{array}
\usepackage[caption=false,font=normalsize,labelfont=sf,textfont=sf]{subfig}
\usepackage{textcomp}
\usepackage{stfloats}
\usepackage{url}
\usepackage{verbatim}
\usepackage{graphicx}
\usepackage{cite}
\usepackage{color}
\usepackage{hyperref}
\hypersetup{
colorlinks=true,
urlcolor=black,
citecolor=blue,
linkcolor=blue,
}

\newcommand{\eg}{\textit{e.g.}}
\newcommand{\ie}{\textit{i.e.}}

\def\METHODNAME{FRESCO}

\begin{document}

\title{Zero-Shot Video Translation and Editing with Frame Spatial-Temporal Correspondence}

\author{Shuai~Yang,~\IEEEmembership{Member,~IEEE},
        Junxin Lin,
        Yifan Zhou,\\
        Ziwei Liu,~\IEEEmembership{Member,~IEEE},
        and~Chen Change Loy,~\IEEEmembership{Senior Member,~IEEE}\vspace{-4mm}
\thanks{Shuai Yang and Junxin Lin are with Wangxuan Institute of Computer Technology, Peking University, Beijing 100080, China.
(E-mail: williamyang@pku.edu.cn, ljx15950821918@stu.pku.edu.cn).}
\thanks{Yifan Zhou, Ziwei Liu and Chen Change Loy are with S-Lab,
Nanyang Technological University, Singapore 639798.
(E-mail: \{yifan006, ziwei.liu, ccloy\}@ntu.edu.sg).}
}

\markboth{ }
{Shell \MakeLowercase{\textit{et al.}}: A Sample Article Using IEEEtran.cls for IEEE Journals}

\maketitle

\begin{abstract}
The remarkable success in text-to-image diffusion models has motivated extensive investigation of their potential for video applications. 
Zero-shot techniques aim to adapt image diffusion models for videos without requiring further model training. Recent methods largely emphasize integrating inter-frame correspondence into attention mechanisms. 
However, the soft constraint applied to identify the valid features to attend is insufficient, which could lead to temporal inconsistency. 
In this paper, we present FRESCO, which integrates intra-frame correspondence with inter-frame correspondence to formulate a more robust spatial-temporal constraint. 
This enhancement ensures a consistent transformation of semantically similar content between frames. Our method goes beyond attention guidance to explicitly optimize features, achieving high spatial-temporal consistency with the input video, significantly enhancing the visual coherence of manipulated videos. We verify FRESCO adaptations on two zero-shot tasks of video-to-video translation and text-guided video editing.
Comprehensive experiments demonstrate the effectiveness of our framework in generating high-quality, coherent videos, highlighting a significant advance over current zero-shot methods. Code: \url{https://github.com/Sunnycookies/FRESCO-v2}.
\end{abstract}


\section{Introduction}
\label{sec:intro}

\IEEEPARstart{I}{n} the current digital era, short videos have become a leading entertainment medium. Editing and creatively rendering these videos is of considerable practical importance. Recent progress in diffusion models~\cite{rombach2022high,ramesh2022hierarchical,saharia2022photorealistic} has significantly revolutionized image editing and style transfer by allowing users to easily manipulate images using natural language prompts. 
However, despite these advances in image manipulation, video manipulation remains challenging, particularly in maintaining temporal consistency in edited videos.

The researchers have proposed three approaches to adapt image models to video domain based on different training requirements, ranging from high to low. 
Data-driven approaches train extended image models on extensive video datasets~\cite{ho2022imagen,singer2022make,esser2023structure}, which requires very high computation cost and a huge amount of data.
One-shot approaches finetune refactored image models on a single video~\cite{wu2023tune,shin2023edit,liu2023video}, which is inconvenient during inference.
Alternatively, zero-shot approaches~\cite{qi2023fatezero,wang2023zero,ceylan2023pix2video,geyer2023tokenflow,cong2023flatten,khachatryan2023text2video,yang2023rerender} offer an efficient avenue for video manipulation by altering the inference process of image models with extra temporal consistency constraints.

\begin{figure}[t]
\centering
\includegraphics[width=0.98\linewidth]{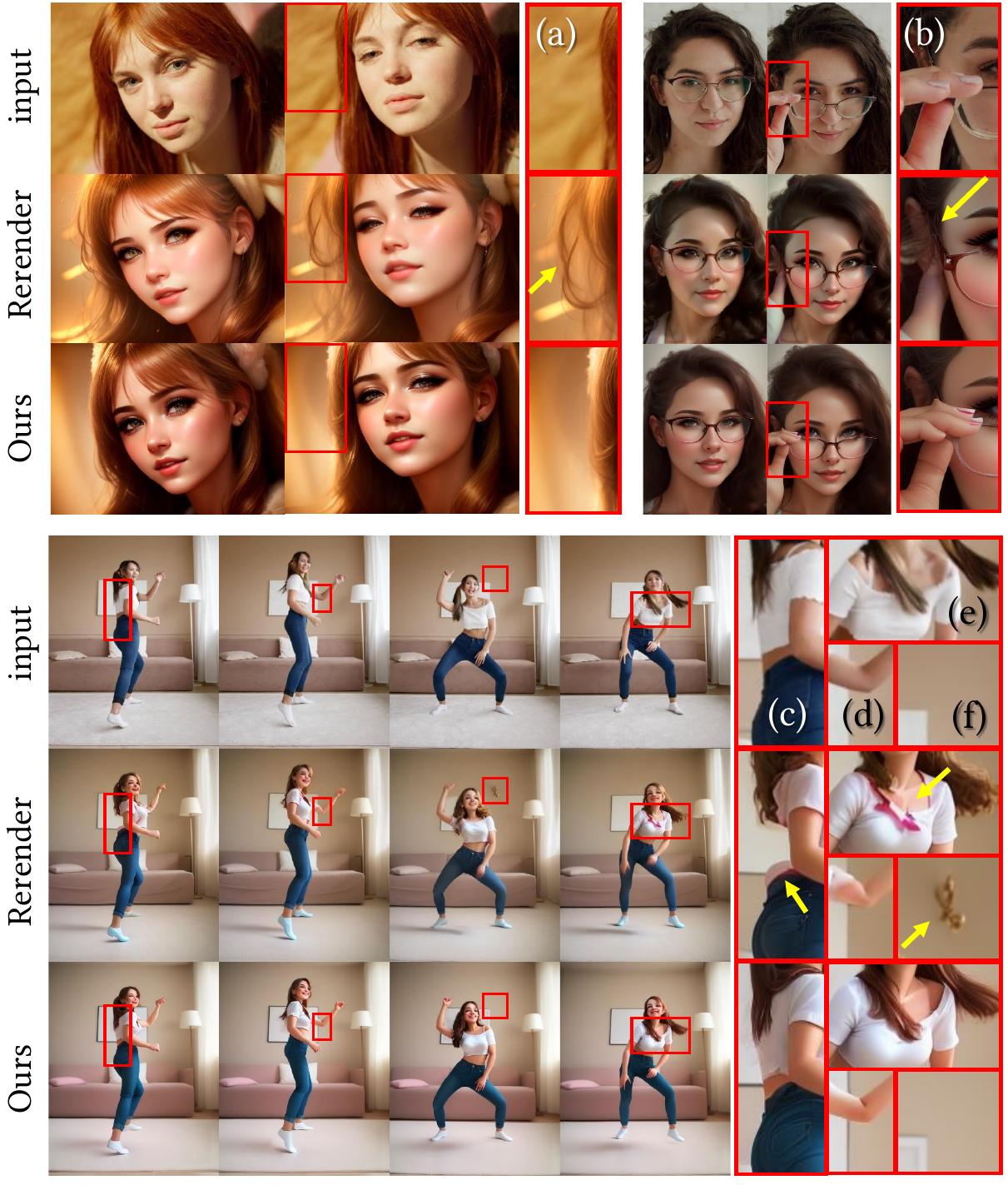}\vspace{-2mm}
\caption{Real video to CG video translation. Methods~\cite{yang2023rerender} relying on optical flow alone suffer (a)(f) inconsistent or (c)(d)(e) missing optical flow guidance and (b) error accumulation. By introducing \METHODNAME, our method addresses these challenges well.}\vspace{-5mm}
\label{fig:challenge}
\end{figure}

The predominant focus of current zero-shot video manipulation methods revolves around attention mechanisms, where cross-frame attention~\cite{wu2023tune, khachatryan2023text2video} is introduced to aggregate features from multiple frames to improve coarse-level global style consistency. 
To enhance fine-grained temporal consistency, methods like Rerender-A-Video~\cite{yang2023rerender} and FLATTEN~\cite{cong2023flatten} 
introduce the optical flow from the input video to enforce inter-frame correspondence.
Although promising, three challenges remain unsolved.
\textbf{1) Inconsistency.} Changes in optical flow during manipulation can cause inconsistent guidance, where foreground elements appear in static background areas without moving properly (Figs.~\ref{fig:challenge}(a)(f)). 
\textbf{2) Undercoverage.} Occlusion or rapid motion can impair the accuracy of optical flow estimation, leading to inadequate constraints and distortions (Figs.~\ref{fig:challenge}(c)-(e)). 
\textbf{3) Inaccuracy.} Frame-by-frame generation may accumulate errors over time, leading to missing fingers in Fig.~\ref{fig:challenge}(b) due to lack of reference in earlier frames.

To address the above critical issues, we present FRamE Spatial-temporal COrrespondence (\textbf{\METHODNAME}). 
Earlier techniques mainly emphasize maintaining \textit{inter-frame temporal correspondence}. 
In this paper, we argue that maintaining \textit{intra-frame spatial correspondence} is equally vital. 
Our method ensures that similar content within a single frame is cohesively manipulated, retaining its semantic similarity after manipulation. 
This strategy effectively tackles the first two challenges:
It prevents the foreground from being mistakenly translated into the background and improves the consistency of optical flow. Regions where optical flow is not available can be regulated by the spatial correspondence of the original frame as in Fig.~\ref{fig:challenge}.

\begin{figure*}[t]
\centering
\includegraphics[width=\linewidth]{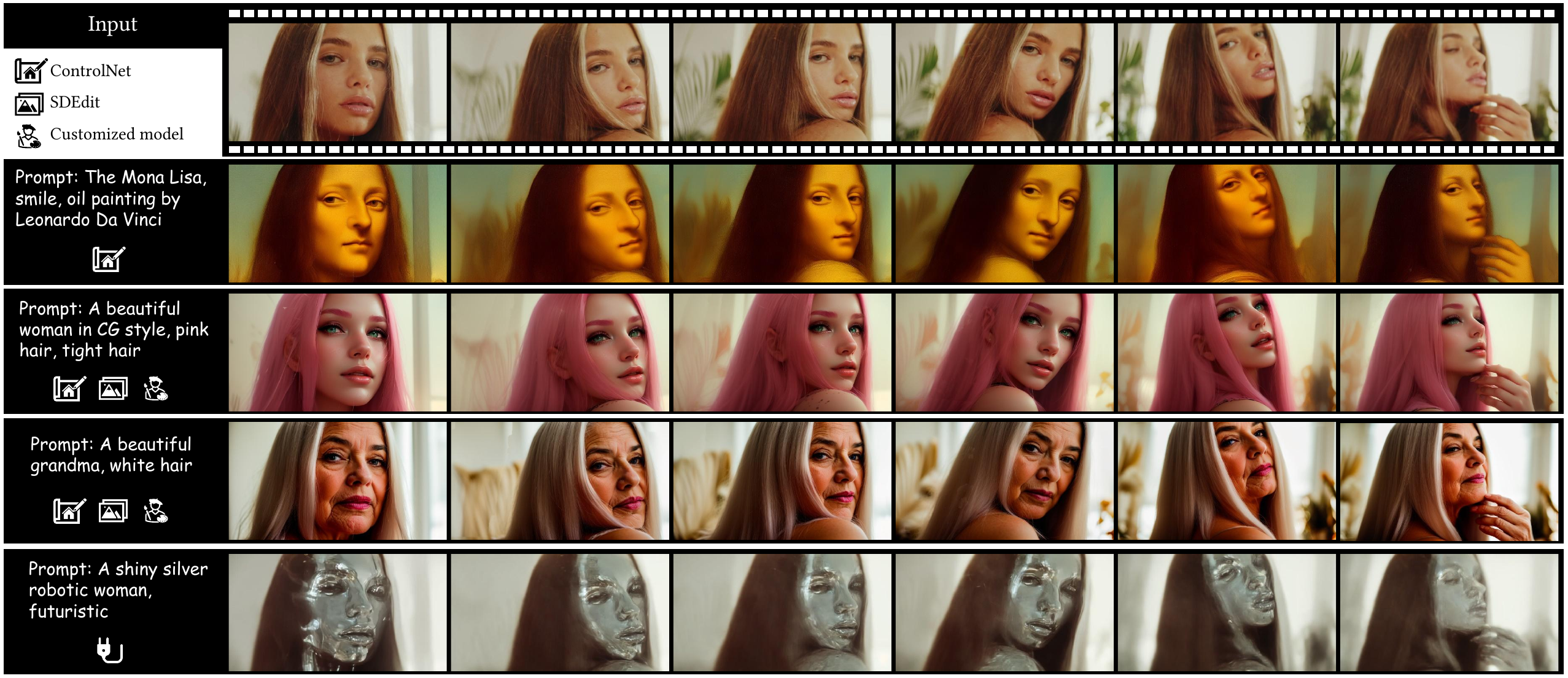}\vspace{-2mm}
\caption{Our framework enables high-quality and coherent video translation based on pre-trained image diffusion model. Given an input video, our method re-renders it based on a target text prompt, while preserving its semantic content and motion. Our zero-shot framework is compatible with various techniques like ControlNet, SDEdit, Play-and-Play and LoRA, enabling flexible translation and editing.}\vspace{-2mm}
\label{fig:teaser}
\end{figure*}

Methodologically, we introduce \METHODNAME~at two levels of attention and feature.
At the attention level, we present \METHODNAME-guided attention, which utilizes not only optical flow guidance~\cite{cong2023flatten} but also the self-similarity of the input frame. This allows for the effective integration of both inter-frame and intra-frame cues from the video input, effectively guiding attention to the most relevant features. 
At the feature level, we introduce \METHODNAME-aware feature optimization. Beyond merely guiding feature attention implicitly, we explicitly update the semantically meaningful features in the U-Net decoder layers to achieve high spatial-temporal consistency with the input video, using gradient descent. 
The combination of these two levels notably improves performance, as demonstrated in Fig.~\ref{fig:teaser}. 
To tackle the final challenge, we employ a hybrid framework. Frames within a batch are jointly processed to guide each other, while anchor frames are shared across batches to maintain inter-batch consistency. Long video manipulation are realized through keyframe selection, manipulation and non-keyframe interpolation.

Compared with our previous work~\cite{yang2024fresco}, we further instantiate our \METHODNAME~adaptation on more image manipulation techniques. Our conference version mainly explores adapting image-conditioned generation technique of ControlNet~\cite{zhang2023adding} and SDEdit~\cite{meng2021sdedit} with \METHODNAME~for video-to-video translation. In this paper, we further consider another important zero-shot image manipulation task: text-guided image editing. We adapt representative Plug-and-Play~\cite{tumanyan2023plug} with \METHODNAME~for text-guided video editing. Moreover, we instantiate the proposed \METHODNAME~adaptation on two hybrid frameworks based on video interpolation techniques of Ebsynth~\cite{jamrivska2019stylizing} and TokenFlow~\cite{geyer2023tokenflow} to facilitate long video manipulation. In addition, comprehensive experiments are conducted to analyze the effectiveness of \METHODNAME~to improve the performance of the above various techniques on videos.
In summary, our main contributions are:
\begin{itemize}
    \item A novel zero-shot diffusion framework guided by frame spatial-temporal correspondence for coherent and flexible video manipulation.
    \item Integrate \METHODNAME-guided attention and \METHODNAME-aware feature optimization as a robust intra-and inter-frame constraint, offering improved consistency and coverage than optical flow alone.
    \item An effective hybrid framework for long video manipulation by jointly processing batched frames with inter-batch consistency.
    \item Instantiation of \METHODNAME~adaptation~on existing image manipulation techniques for text-guided video translation and video editing.
\end{itemize}

The rest of this paper is organized as follows.
In Section~\ref{sec:related_work}, we review related studies in image diffusion models and zero-shot text-guided video manipulation.
In Section~\ref{sec:method}, preliminary text-guided image manipulation models and the details of the proposed spatial-temporal correspondence to adapt image models to video models are elaborated.
Section~\ref{sec:instantiation} instantiates the proposed \METHODNAME~adaptation on zero-shot video translation and video editing for keyframe manipulation, as well as non-keyframe interpolation for long video manipulation. 
Section~\ref{sec:experiment} validates the superiority of our method via extensive experiments and comparisons with state-of-the-art zero-shot methods.
Finally, we conclude our work in Section~\ref{sec:conclusion}.

\section{Related Work}
\label{sec:related_work}

\subsection{Image Diffusion Models}
\label{sec:2.1}

In recent years, there has been a rapid expansion of image diffusion models~\cite{ramesh2022hierarchical,saharia2022photorealistic} for text-guided image generation and editing. These models create images through an iterative denoising process~\cite{ho2020denoising}.
Stable Diffusion builds on the latent diffusion model~\cite{rombach2022high}, performing denoising with a U-Net network in a condensed latent space to decrease complexity.
Recent practices~\cite{esser2024scaling,chenpixart,chen2024pixart,chen2025pixart} use DiT\cite{peebles2023scalable} as the denoising backbone, which guarantees scalability, efficiency, and high-quality sample generation. In addition to general image generation, object appearances and styles can be further customized by finetuning text embeddings~\cite{gal2022image}, model weights~\cite{ruiz2023dreambooth,kumari2022customdiffusion,han2023svdiff,hulora} or encoders~\cite{wei2023elite,xu2023prompt,ye2023ip, gong2023talecrafter, gal2023encoder}.

Text-to-image models have led to the development of various image manipulation frameworks~\cite{hertz2022prompt,brooks2023instructpix2pix},
among which the most relevant to our task are \textbf{1)} image-conditioned generation and \textbf{2)} text-guided image editing.
Image-conditioned generation, or image-to-image translation, introduces extra image conditions to specify the spatial characteristics of the generated image.
SDEdit~\cite{meng2021sdedit} implements image guidance in the generation process.
ControlNet~\cite{zhang2023adding}, T2I-Adapter~\cite{mou2024t2i} and ControlNeXt~\cite{peng2024controlnext} integrate a control path to provide structure or layout information for fine-grained generation.
Text-guided image editing aims to modify the content of an image with an editing prompt.
Prompt2Prompt~\cite{hertz2022prompt} and Plug-and-Play~\cite{tumanyan2023plug} employ cross-attention control or self-attention injection to maintain the image layout during editing.
To edit real images, DDIM inversion~\cite{songdenoising} and Null-Text Inversion~\cite{mokady2023null} have been proposed to embed real images in the noisy latent feature, facilitating editing with attention modulation~\cite{parmar2023zero,tumanyan2023plug,cao2023masactrl}.

Our zero-shot framework does not alter the pre-trained model and therefore is compatible with the aforementioned image manipulation mechanism for flexible control and customization as shown in Fig.~\ref{fig:teaser}.

\subsection{Zero-Shot Text-Guided Video Manipulation}

Although large video diffusion models trained or fine-tuned on videos have been studied~\cite{ho2022imagen,singer2022make,he2022latent,zhou2022magicvideo,luo2023videofusion,esser2023structure,blattmann2023align,guo2023animatediff,wu2023tune,shin2023edit,feng2023ccedit,ge2023pyoco,wang2023lavie,yang2024cogvideox,jin2024pyramidal,gu2025diffusion}, this paper focuses on lightweight and highly compatible zero-shot methods, especially those built upon U-Net-based Stable Diffusion models. 
Zero-shot methods mainly originated from extending existing image manipulation  framework to the video domain. 
Based on image manipulation tasks, they can be broadly classified into two types:
\textbf{1)} video-to-video translation and \textbf{2)} text-guided video editing methods.

Zero-shot video-to-video translation is based on zero-shot image-conditioned  generation techniques such as ControlNet~\cite{zhang2023adding} and SDEdit~\cite{meng2021sdedit}.
Text2Video-Zero~\cite{khachatryan2023text2video} simulates motions by modulating noise. ControlVideo~\cite{zhang2023controlvideo} extends ControlNet to videos by incorporating cross-frame attention and frame interpolation.
VideoControlNet~\cite{hu2023videocontrolnet} and Rerender-A-Video~\cite{yang2023rerender} achieve enhanced temporal coherence by warping and blending previously translated frames using optical flow.

Zero-shot text-guided video editing stem from zero-shot image editing techniques such as Prompt-to-Prompt~\cite{hertz2022prompt} and Plug-and-Play~\cite{tumanyan2023plug}. These methods~\cite{qi2023fatezero,jeong2023ground} apply DDIM inversion to video frames and record attention features to facilitate attention control during the editing of designated content.
FateZero~\cite{qi2023fatezero} identifies and preserves the unedited region and employs cross-frame attention to enforce global appearance coherence.
Pix2Video~\cite{ceylan2023pix2video} updates the latent of the current frame guided by the latent of the previous frame to suppress flickers. 
To explicitly leverage inter-frame correspondence, TokenFlow~\cite{geyer2023tokenflow} matches and blends features from the edited keyframes to interpolate non-keyframes.
FLATTEN~\cite{cong2023flatten} incorporates optical flows into the attention mechanism for fine-grained temporal consistency.
VidToMe~\cite{li2024vidtome} matches and merges tokens so that matched tokens use the same information during attention, leading to more consistent results with decreased computational cost.

Video editing methods employ DDIM inversion features to preserve the details of the original frames, making them particularly effective for localized editing. Meanwhile, video translation methods globally re-render the entire frames, offering greater creative flexibility. However, without DDIM inversion features for guidance, video translation methods are more susceptible to flickering. We experimentally find that these two methods exhibit distinct characteristics with significantly different outcomes on the same input (please refer to our supplementary material).  In this work, we incorporate our \METHODNAME~into both video manipulation frameworks to enhance temporal consistency while maintaining each framework's unique attributes to suit various tasks.

\begin{figure*}[t]
\centering
\includegraphics[width=\linewidth]{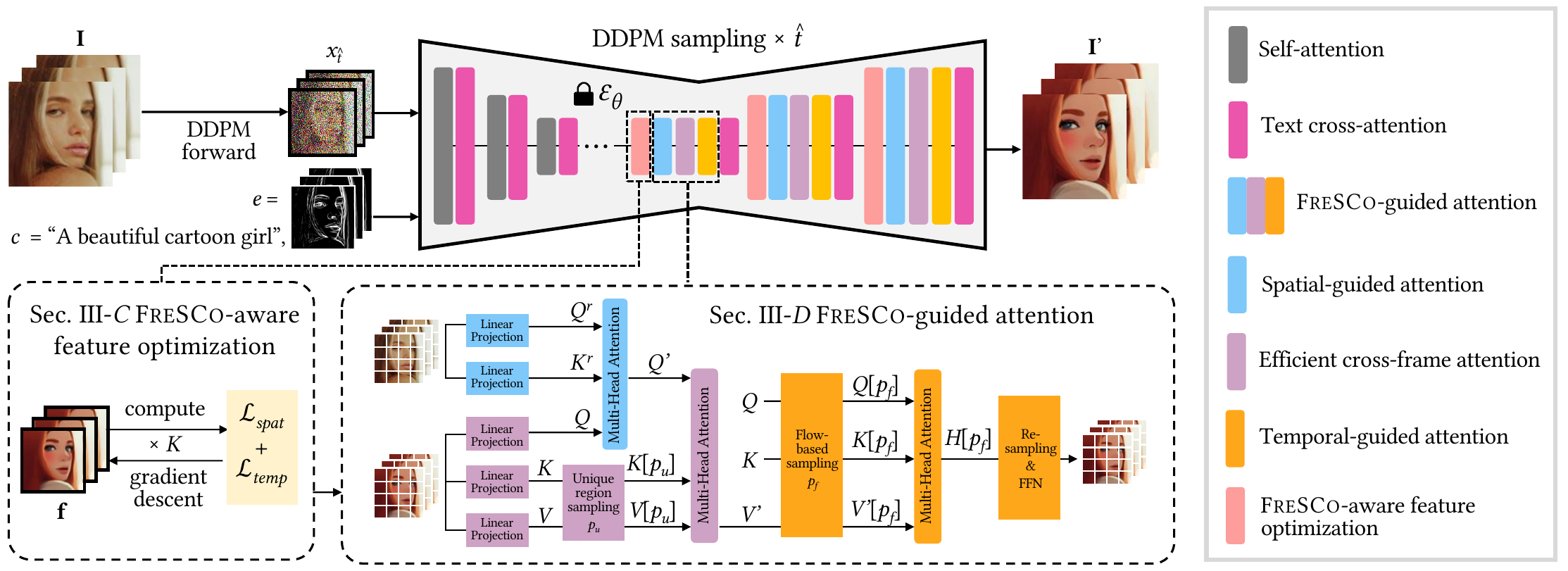}\vspace{-2mm}
\caption{Framework of our zero-shot video translation guided by FRamE Spatial-temporal COrrespondence (\METHODNAME). A \METHODNAME-aware optimization is applied to the U-Net features to strengthen their temporal and spatial coherence with the input frames. We integrate \METHODNAME~into self-attention layers, resulting in spatial-guided attention to keep spatial correspondence with the input frames, efficient cross-frame attention and temporal-guided attention to keep rough and fine temporal correspondence with the input frames, respectively.}\vspace{-3mm}
\label{fig:pipeline}
\end{figure*}

\section{Methodology}
\label{sec:method}

\subsection{Preliminary}
\label{sec:preliminary}

\textbf{Stable Diffusion.} Stable Diffusion contains a forward diffusion process and a backward sampling process in the latent space.
An input frame $I$ is first mapped to a latent feature $x_0=\mathcal{E}(I)$ with an Encoder $\mathcal{E}$.
The DDPM forward process~\cite{ho2020denoising} progressively adds Gaussian noise to $x_0$ until the time step $T$. At the step $t$, the noisy latent is obtained as  
\begin{equation}\label{eq:forward_sample}
  x_t=\sqrt{\bar{\alpha}_t}x_0 + \sqrt{1-\bar{\alpha}_t}\epsilon,
\end{equation}
where $\bar{\alpha}_t$ is a pre-defined hyperparamter at the step $t$ and $\epsilon$ is a randomly sampled standard Gaussian noise. 
The backward sampling process~\cite{ho2020denoising} samples an image by iteratively denoising a random Gaussian noise $x_T\sim\mathcal{N}(0, \mathbf{I})$ with a trained U-Net $\epsilon_\theta$. 
We focus on two best-known sampling schemes: DDPM and DDIM samplings. DDPM sampling predicts $x_{t-1}$ from $x_{t}$ by 
\begin{equation}\label{eq:ddpm}
  x_{t-1}=\frac{\sqrt{\bar\alpha_{t-1}}\beta_t }{1-\bar\alpha_t}\hat{x}_0 + \frac{(1-\bar\alpha_{t-1})(\sqrt{\alpha_t}x_t+\beta_t\epsilon)}{1-\bar\alpha_t},
\end{equation}
where $\beta_t$ is a pre-defined hyperparamter and $\hat{x}_0$ is the predicted $x_0$ at the denoising step $t$,
\begin{equation}\label{eq:denoise}
  \hat{x}_0=(x_t-\sqrt{1-\bar\alpha_t}\epsilon_\theta(x_t, t, c))/\sqrt{\bar\alpha_t},
\end{equation}
and $\epsilon_\theta(x_t, t, c)$ is the predicted noise of $x_t$ based on the step $t$, the text prompt $c$.
Alternatively, DDIM provides a deterministic sampling scheme
\begin{equation}\label{eq:ddim}
  x_{t-1}=\sqrt{\bar\alpha_{t-1}}{\hat{x}_0}+\sqrt{1-\bar\alpha_{t-1}}\epsilon_\theta(x_t, t, c).
\end{equation}
The final image $I=\mathcal{D}(x_0)$ is generated with a Decoder $\mathcal{D}$.

\textbf{Zero-shot image-conditioned generation.} 
SDEdit~\cite{meng2021sdedit} incorporate an image $I$ as extra conditions by sampling images from a noisy  $x_{\hat{t}}=\sqrt{\bar{\alpha}_{\hat{t}}}x_0+\sqrt{1-\bar{\alpha}_{\hat{t}}}\epsilon$ at an intermediate step $\hat{t}$. 
SDEdit iteratively translates $x'_{\hat{t}}=x_{\hat{t}}$ into $x'_0$ following Eq.~(\ref{eq:ddpm}) or Eq.~(\ref{eq:ddim}) to obtain the final result $I'$.
Meanwhile, ControlNet~\cite{zhang2023adding} can be used to provide additional structure or layout information $e$ (\eg, edge maps extracted from $I$) to guide the sampling.  
Specifically, a ControlNet branch is added to the U-Net as $\epsilon_\theta(x'_t, t, c, e)$.

\textbf{Zero-shot text-guided image editing.} Zero-shot editing methods first apply DDIM inversion~\cite{songdenoising} to convert the image $I$ to be edited to the noisy latent $x_T, \{\phi_t\}=\text{DDIM\_inv}(\mathcal{E}(I))$, where $\phi_t$ is the intermediate parameters stored during the DDIM inversion process at step $t$. Then, guided by the target prompt $c$, DDIM sampling is applied to gradually denoise $x'_T=x_T$ to $x'_0$ with the intermediate features injected as $\epsilon_\theta(x'_t, t, c, \phi_t)$ in Eq.~(\ref{eq:ddim}).
Specifically, Plug-and-Play~\cite{tumanyan2023plug} replaces the residual feature and self-attention query and key vectors of the U-Net with those stored from the DDIM inversion process of $I$ to improve structure consistency.

For simplicity, we will omit the denoising step $t$ in the following. Details of Preliminary are provided in the supplementary pdf.

\subsection{\textbf{\METHODNAME} Overview}

For a series of video frames $\mathbf{I}=\{I_i\}^N_{i=1}$, we follow the text-guided image editing or image-conditioned generation pipeline outlined in Sec.~\ref{sec:preliminary} to manipulate $\mathbf{I}$ to obtain the corresponding $\mathbf{I}'=\{I'_i\}^N_{i=1}$. Our adaptation emphasizes the integration of the spatial and temporal correspondences of $\mathbf{I}$ into the U-Net. In particular, we specify the temporal and spatial correspondences of $\mathbf{I}$ as:
\begin{itemize}
    \item \textbf{Temporal correspondence}. We model the inter-frame correspondence via optical flows between frames to maintain temporal consistency. Let $w^j_i$ and $M^j_i$ denote the optical flow and occlusion mask from $I_i$ to $I_j$, respectively. Our goal is to ensure that $I'_i$ and $I'_{i+1}$ share $w^{i+1}_i$ in non-occluded areas.
    \item \textbf{Spatial correspondence}. The intra-frame correspondence refers to the similarity within a frame, measured by the self-similarity among pixels. The objective of $I'_i$ is to match the self-similarity of $I_i$, \ie, semantically similar elements should appear similarly after manipulation, and vice versa. 
    This preservation of semantic and spatial structure implicitly enhances the temporal consistency.
\end{itemize}

We concentrate our adaptation on the \textit{input feature} and \textit{attention module} of the semantical meaningful and less noisy decoder layer in the U-Net (will discussed later in Fig.~\ref{fig:ablation_optimize}):
\begin{itemize}
    \item \textbf{Feature adaptation}.~We introduce a novel \METHODNAME-aware feature optimization method, depicted in Fig.~\ref{fig:pipeline}. We design a spatial consistency loss $\mathcal{L}_{spat}$ together with a temporal consistency loss $\mathcal{L}_{temp}$ to directly refine the decoder-layer features $\mathbf{f}=\{f_i\}_{i=1}^N$, enhancing their spatial and temporal coherence with the input frames.
    \item \textbf{Attention adaptation}.~We substitute self-attentions with \METHODNAME-guided attentions, consisting of three components as shown in Fig.~\ref{fig:pipeline}: a spatial-guided attention to aggregate features based on the self-similarity of the input frame; a cross-frame attention to gather features from all frames; a temporal-guided attention to aggregate features along the same optical flow to  boost temporal consistency.
\end{itemize}

The proposed feature adaptation method enhances coherence by directly refining the feature for increased spatial and temporal consistency with $\mathbf{I}$. At the same time, our attention adaptation method enhances coherence indirectly through soft constraints that determine how and where to attend to valid features. 
Our results show that the integration of both adaptation techniques yields the best performance.

\subsection{\textbf{\METHODNAME}-Aware Feature Optimization}
\label{sec:feature}

The input feature $\mathbf{f}=\{f_i\}_{i=1}^N$ of each decoder layer of U-Net
is updated by gradient descent through optimizing
\begin{equation}
  \hat{\mathbf{f}}=\arg\min_{\mathbf{f}} \mathcal{L}_{temp}(\mathbf{f}) + \mathcal{L}_{spat}(\mathbf{f}).
\end{equation}
The updated $\hat{\mathbf{f}}$ replaces $\mathbf{f}$ for subsequent processing.

For the temporal consistency loss $\mathcal{L}_{temp}$, we aim for the feature values at corresponding positions in every pair of consecutive frames to remain consistent,
\begin{equation}\label{eq:temporal_loss}
  \mathcal{L}_{temp}(\mathbf{f}) = \sum_i\|M_i^{i+1}(f_{i+1}-w_i^{i+1}(f_i))\|_1
\end{equation}

For the spatial consistency loss $\mathcal{L}_{spat}$, we employ cosine similarity in the feature space to evaluate the spatial correspondence of $I_i$. 
Let $f^r_i$ denote the extracted U-Net decoder feature of $I_i$ (we will detail the feature extraction method in Sec.~\ref{sec:instantiation}).
Let $\tilde{f}$ represent the normalized $f$, ensuring that each component of $\tilde{f}$ is a unit vector. 
The cosine similarity of all pair of elements can be simply computed as the gram matrix~\cite{gatys2016image} of the normalized feature.
We aim for the gram matrix of $\tilde{f}_i$ to approach the gram matrix of $\tilde{f}^r_i$,
\begin{equation}\label{eq:content_loss}
  \mathcal{L}_{spat}(\mathbf{f})= \lambda_\text{spat}\sum_i\|\tilde{f}_i\tilde{f}_i^{~\top}-\tilde{f}^r_i\tilde{f}_i^{r\top}\|^2_2.
\end{equation}

\subsection{\textbf{\METHODNAME}-Guided Attention}
\label{sec:attention}

\begin{figure}[t]
\centering
\includegraphics[width=\linewidth]{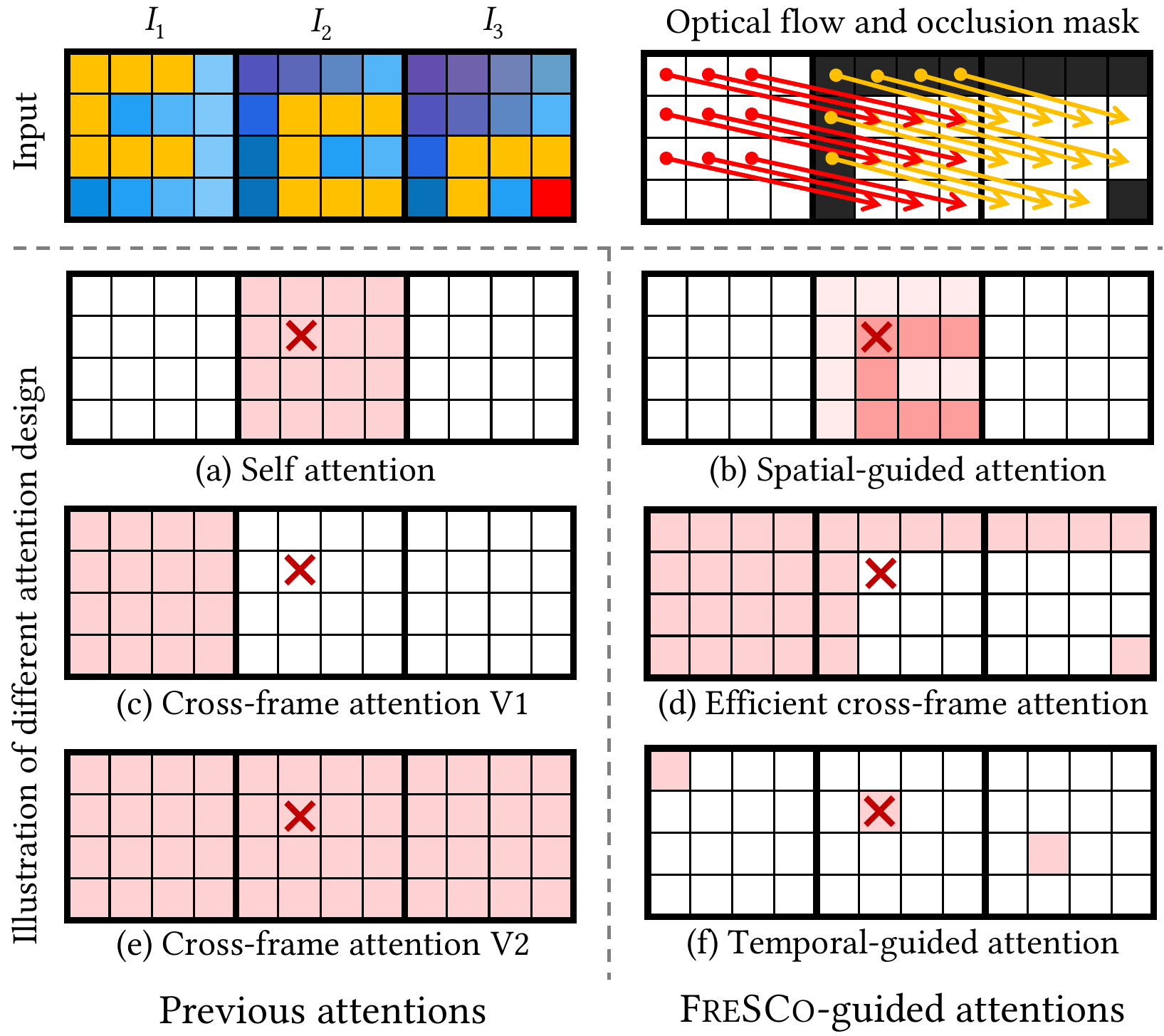}\vspace{-2mm}
\caption{Illustration of attention mechanism. The patches marked with red crosses attend to the colored patches and aggregate their features. Compared to previous attentions, \METHODNAME-guided attention further considers intra-frame and inter-frame correspondences of the input. (b) Spatial-guided attention aggregates intra-frame features based on the self-similarity of the input frame (darker indicates higher weights). (d) Efficient cross-frame attention eliminates redundant patches and retains unique patches. (f) Temporal-guided attention aggregates inter-frame features on the same flow.}\vspace{-4mm}
\label{fig:attention}
\end{figure}

\textbf{Spatial-guided attention.}~Unlike self-attention, spatial-guided attention aggregates patches based on the similarity between tokens before translation rather than their own similarity. 
In particular, we first extract the self-attention query vector $Q^r_i$ and key vector $K^r_i$ of $I_i$  (we will detail the vector extraction method in Sec.~\ref{sec:instantiation}). 
Subsequently, spatial-guided attention aggregates $Q_i$ with
\begin{equation}
\label{eq:sg_attention}
Q'_i=\textit{Softmax}(\frac{Q^r_iK_i^{r\top}}{\lambda_s\sqrt{d}})\cdot Q_i,
\end{equation}
where $\lambda_s$ is a scale factor and $d$ is the query vector dimension. As illustrated in Fig.~\ref{fig:attention}, the foreground tokens mainly aggregate features in the C-shaped foreground region, with reduced attention paid to the background. Thus, $Q'$ shows improved spatial consistency with the input frame compared to $Q$.

\textbf{Efficient cross-frame attention.}~We employ cross-frame attention to maintain global style consistency. Instead of referencing the first frame or the previous frame~\cite{khachatryan2023text2video,ceylan2023pix2video} (Fig.~\ref{fig:attention}(c)), which fails to handle the newly emerged fingers in Fig.~\ref{fig:challenge}(b), or referencing all available frames (Fig.~\ref{fig:attention}(e)), which is computationally intensive, we aim to consider all frames simultaneously with minimal redundancy. To achieve this, we introduce efficient cross-frame attention: except for the first frame, we focus only on regions in each frame that were not visible in its previous frame (\ie, the occlusion region). Thus, we create a unique index $p_u$ for all tokens within these unique areas. The keys and values of these tokens can be extracted as $K[p_u]$, $V[p_u]$, to which cross-frame attention is applied
\begin{equation}
\label{eq:ecf_attention}
V'_i=\textit{Softmax}(\frac{Q'_i(K[p_u])^\top}{\sqrt{d}})\cdot V[p_u].
\end{equation}

\textbf{Temporal-guided attention.}~Inspired by FLATTEN~\cite{cong2023flatten}, we employ flow-based attention to ensure fine-grained cross-frame consistency. As illustrated in Fig.~\ref{fig:attention}, we trace the same tokens across various frames. For each optical flow, we create a flow index $p_f$ that includes all tokens along this flow. In FLATTEN, patches are restricted to attend only to those in other frames, which can lead to instability when a flow has a limited number of valid tokens. In contrast, our temporal-guided attention allows attending the current frame's token,
\begin{equation}
\label{eq:tg_attention}
H[p_f]=\textit{Softmax}(\frac{Q[p_f](K[p_f])^\top}{\lambda_t\sqrt{d}})\cdot V'[p_f],
\end{equation}
where $\lambda_t$ is a scale factor. And $H$ is the final output of our \METHODNAME-guided attention layer.

\section{Instantiation}
\label{sec:instantiation}

This section instantiates the proposed \METHODNAME~adaptation on two kinds of representative zero-shot video manipulation frameworks. We first adapt the image translation pipeline of Stable Diffusion based on ControlNet~\cite{zhang2023adding} and SDEdit~\cite{meng2021sdedit} to the video translation in Sec.~\ref{sec:sdedit}.
Then, we apply our adaptation to Plug-and-Play~\cite{tumanyan2023plug} for text-guided video editing in Sec.~\ref{sec:pnp}. 
Finally, for long video manipulation, we explore a hybrid framework with the frame interpolation techniques of EbSynth~\cite{jamrivska2019stylizing} and TokenFlow~\cite{geyer2023tokenflow} in Sec.~\ref{sec:tokenflow}.

\subsection{Zero-Shot Video-to-Video Translation}
\label{sec:sdedit}
 
As elaborated in Sec.~\ref{sec:preliminary}, the input of our video-to-video translation task is the source video frames $\mathbf{I}$, target prompt $c$ and SDEdit parameter timestep $\hat{t}$. 
For each frame $I\in\mathbf{I}$, we first encode it $x_0=\mathcal{E}(I)$  and extract its structure information $e$ (\eg, canny edges) as input to the ControlNet branch. 
Then, we perform a single-step DDPM forward and backward process on $I$ (\ie, add noise to $x_0$ to obtain $x_1$ and then denoise $x_1$ using the original U-Net), and extract the U-Net decoder feature $f^r$ and self-attention query vector $Q^r$ and key vector $K^r$. Since a single-step forward process adds negligible noises, $f^r$, $Q^r$ and $K^r$ can serve as semantic meaningful features of $I$ to calculate its self-similarity.
Furthermore, optical flows and occlusion masks between input frames are estimated with GMFlow~\cite{xu2022gmflow}, which are used to derive the indexes $p_u$ and $p_f$. For simplicity, we denote all \METHODNAME~parameters as $\omega$.

After the above preparation, we can now perform SDEdit to obtain the initial noisy latent $x'_{\hat{t}}=x_{\hat{t}}$ using Eq.~(\ref{eq:forward_sample}), and denoise it with DDPM sampling and \METHODNAME~adaptation,
\begin{align}
  x'_{t-1}&=\frac{\sqrt{\bar\alpha_{t-1}}\beta_t }{1-\bar\alpha_t}\hat{x}'_0 + \frac{(1-\bar\alpha_{t-1})(\sqrt{\alpha_t}x'_t+\beta_t\epsilon)}{1-\bar\alpha_t},\\
  \hat{x}'_0&=(x'_t-\sqrt{1-\bar\alpha_t}\epsilon_\theta(x'_t, t, c, e, \omega))/\sqrt{\bar\alpha_t},
\end{align}
where the self-attention layer of the U-Net $\epsilon_\theta$ is replaced with our proposed \METHODNAME-guided attention layer and the \METHODNAME-aware feature optimization is activated in each decoder layer of the U-Net.
Specifically, the extracted $f^r$, $Q^r$ and $K^r$ are used in the spatial consistency loss $\mathcal{L}_{spat}$ (Eq.~(\ref{eq:content_loss})) and the spatial-guided attention (Eq.~(\ref{eq:sg_attention})), respectively. 
The estimated optical flows and occlusion masks are used to compute the temporal consistency loss $\mathcal{L}_{temp}$ (Eq.~(\ref{eq:temporal_loss})). Their derived $p_u$ and $p_f$ are used in efficient cross-frame attention (Eq.~(\ref{eq:ecf_attention})) and temporal-guided attention (Eq.~(\ref{eq:tg_attention})), respectively. 
Finally, the translated frame $I'=\mathcal{D}(x'_0)$ is obtained and all resulting frames are merged into the translated video $\mathbf{I}'$.

\begin{figure}[t]
\centering
\includegraphics[width=\linewidth]{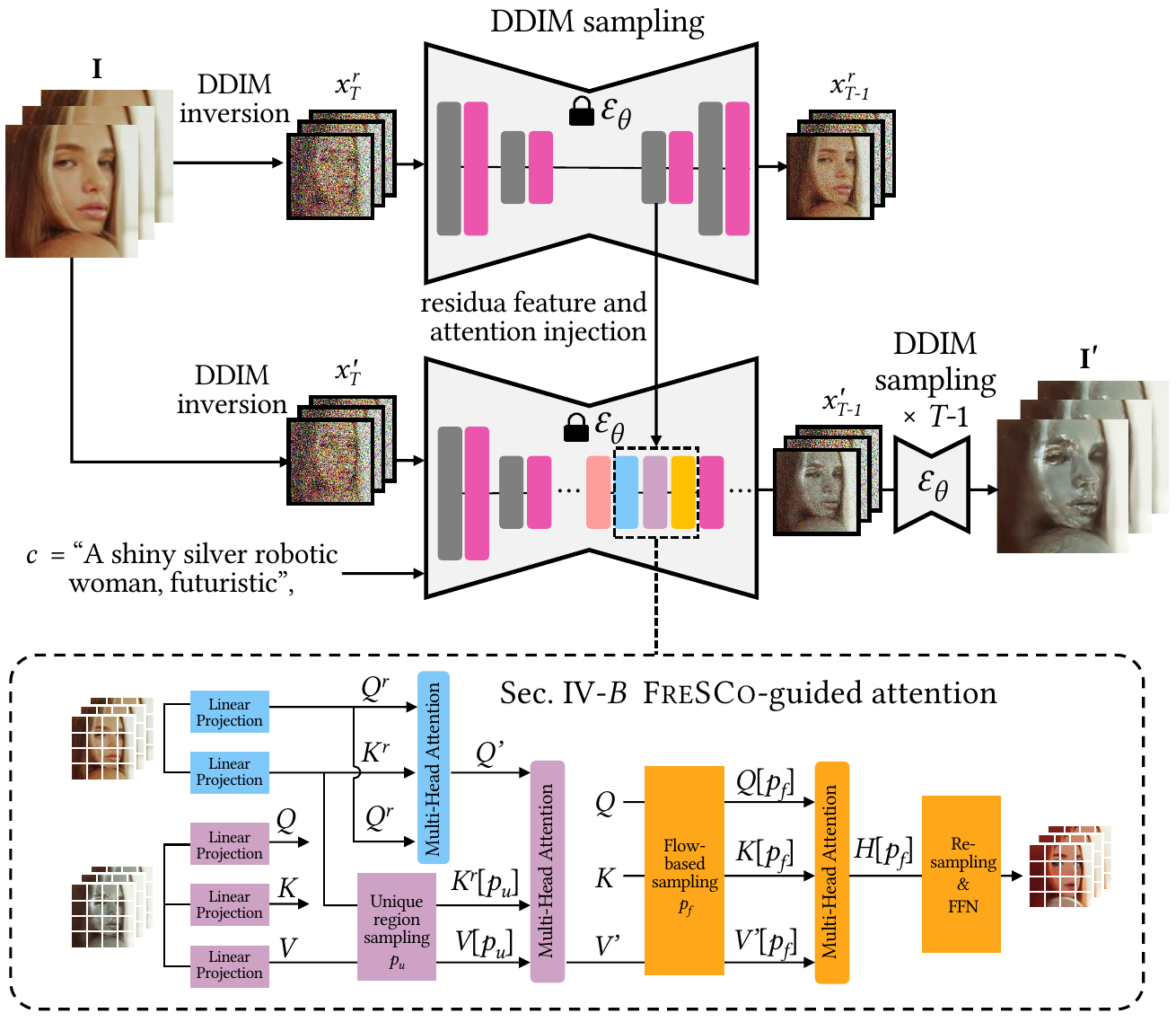}\vspace{-2mm}
\caption{Framework of our zero-shot video editing guided by \METHODNAME.}\vspace{-3mm}
\label{fig:pipeline2}
\end{figure}

\subsection{Zero-Shot Text-Guided Video Editing}
\label{sec:pnp}

Similar to Sec.~\ref{sec:sdedit}, we first prepare the parameters of \METHODNAME, and perform Plug-and-Play to edit frames with the U-Net equipped with \METHODNAME~constraints. Specifically, the input of our text-guided video editing task is the source video frames $\mathbf{I}$ and the target prompt $c$. For each frame $I\in\mathbf{I}$, we first encode it to  $x_0$. Then, we perform DDIM inversion to $x_0$ to obtain the initial noisy latent $x_T$ using the original U-Net, 
\begin{equation}
x_T, \{\phi_t, Q_t^r, K_t^r\} =\text{DDIM\_inv}(x_0),
\end{equation}
where $\phi_t$, $Q_t^r$ and $K_t^r$ are the residual feature, the self-attention query and key vectors of the U-Net of $I$ at the timestep $t$ of DDIM inversion. $\phi_t$ is used for Plug-and-Play, while $Q_t^r$ and $K_t^r$ are for our spatial-guided attention. As in Sec.~\ref{sec:sdedit}, optical flows, occlusion masks, and indexes are prepared.

Next, we perform Plug-and-Play to denoise $x'_T=x_T$ with DDIM sampling and \METHODNAME~adaptation,
\begin{align}
  x'_{t-1}&=\sqrt{\bar\alpha_{t-1}}{\hat{x}'_0}+\sqrt{1-\bar\alpha_{t-1}}\epsilon_\theta(x'_t, t, c, \phi_t, \omega),\\
  \hat{x}'_0&=(x'_t-\sqrt{1-\bar\alpha_t}\epsilon_\theta(x'_t, t, c, \phi_t, \omega))/\sqrt{\bar\alpha_t},
\end{align}
Finally, the translated frame $I'=\mathcal{D}(x'_0)$ is obtained and all resulting frames are merged into the translated video $\mathbf{I}'$. There are two differences from the video translation. 
First, SDEdit adds random noises to obtain $x_{\hat{t}}$ which may harm the inherent self-similarity. In contrast, DDIM inversion better preserves the self-similarity of the frame by adding predicted noises. We find that we can savely remove $\mathcal{L}_{spat}$ when denoising from the DDIM inversion latent $x_T$ without performance degradation. 
Second, Plug-and-Play injects the self-attention query and key vectors $Q^r$ and $K^r$ from the DDIM inversion process into the DDIM sampling process.
We integrate this idea into our spatial-guided attention and efficient cross-frame attention by using $Q^r_i$ and $K^r_i$ instead of $Q_i$ and $K_i$:
\begin{align}
Q'_i&=\textit{Softmax}(\frac{Q^r_iK_i^{r\top}}{\lambda_s\sqrt{d}})\cdot Q^r_i,\\
V'_i&=\textit{Softmax}(\frac{Q'_i(K^r[p_u])^\top}{\sqrt{d}})\cdot V[p_u].
\end{align}

\subsection{Long Video Manipulation}
\label{sec:tokenflow}

The number of frames $N$ that can be processed at one time is limited by GPU memory. 
A practical way of manipulating long videos ($\geq 100$ frames) is to use a hybrid framework~\cite{yang2023rerender,geyer2023tokenflow}: perform video manipulation on keyframes only and interpolate non-keyframes based on manipulated keyframes. In this section, we instantiate the proposed \METHODNAME~adaptation on two hybrid frameworks based on video interpolation techniques of Ebsynth~\cite{jamrivska2019stylizing} and TokenFlow~\cite{geyer2023tokenflow}.

\subsubsection{Hybrid framework with Ebsynth}
\label{sec:ebsynth}

Ebsynth is a traditional frame interpolation technique based on patch matching and Poisson blending~\cite{heitz2018high} in the pixel domain, which is first combined with Stable Diffusion for hybrid zero-shot video translation in Rerender-A-Video~\cite{yang2023rerender}.
We follow its main process of keyframe selection, keyframe manipulation and non-keyframe interpolation.

\textbf{Keyframe selection.} Rerender-A-Video uses uniform keyframe sampling, which is suboptimal. We introduce a heuristic algorithm for keyframe selection, detailed in Algorithm~\ref{alg:algorithm1}. This method relaxes the fixed sampling step to a range $[s_\text{min}, s_\text{max}]$, and increases the density of keyframe sampling when there are significant motions, as assessed by the $L_2$ distance between frames.

\setlength{\textfloatsep}{8pt}
\begin{algorithm}[t]
  \caption{Keyframe selection}
  \textbf{Input:} Video $\mathbf{I}=\{I_i\}^M_{i=1}$, sample parameters $s_\text{min}$, $s_\text{max}$\\
  \textbf{Output:} Keyframe index list $\Omega$ in ascending order
  \begin{algorithmic}[1]
    \State initialize $\Omega=[1, M]$ and $d_i=0, \forall i\in[1,M]$
    \State set $d_i=L_2(I_i,I_{i-1}), \forall i\in[s_\text{min}+1,N-s_\text{min}]$
    \While {\textbf{exists} $i$ \textbf{such that} $\Omega[i+1]-\Omega[i]>s_\text{max}$}
        \State $\Omega.$\texttt{insert}$(\hat{i}).$\texttt{sort}$()$ with $\hat{i}=\arg\max_i(d_i)$
        \State set $d_j=0$, $\forall~j\in(\hat{i}-s_\text{min}, \hat{i}+s_\text{min})$
    \EndWhile
  \end{algorithmic}
  \label{alg:algorithm1}
\end{algorithm}

\textbf{Keyframe manipulation.} Keyframes numbered over $N$ are divided into multiple $N$-frame batches. 
\METHODNAME-guided video manipulation in Sec.~\ref{sec:sdedit} and Sec.~\ref{sec:pnp} is applied per batch.
To impose inter-batch consistency, each batch includes the first and last frames as anchor frames in the previous batch. Specifically, for the $k$-th batch, the keyframe indexes are $\{1,(k-1)(N-2)+2,(k-1)(N-2)+3,...,k(N-2)+2\}$. Furthermore, during all denoising steps, we record the latent features $x'_t$ (Eq.~(\ref{eq:ddpm})) of both the first and last frames for each batch. Subsequently, these stored features are used to substitute the corresponding latent features in the subsequent batch.

\textbf{Non-keyframe interpolation.}~Since Ebsynth functions independently of the Diffusion model, we use the standard Ebsynth without adaptation. 
In particular, it uses pairs of keyframes $(I'_{\Omega[i]}, I'_{\Omega[i+1]})$ to interpolate the non-keyframes that lie between the indexes $\Omega[i]+1$ and $\Omega[i+1]-1$.

\subsubsection{Hybrid framework with TokenFlow}

TokenFlow matches tokens between original keyframes and non-keyframes to obtain a nearest-neighbor field, and blends tokens of the edited keyframes to interpolate features of the edited non-keyframes based on this nearest-neighbor field. Hybrid framework with TokenFlow also contains keyframe selection, keyframe manipulation and non-keyframe interpolation.

\textbf{Keyframe selection.} TokenFlow handles keyframes and non-keyframes iteratively. In each timestep, keyframes are randomly sampled and edited, followed by non-keyframe interpolation. This random sampling strategy prevents an obvious visual discrepancy between keyframes and other frames. However, since every frame could be assigned as keyframes at any timestep, we will need to store all possible \METHODNAME~parameters of the entire video, which are memory-inefficient for inter-frame parameters. 
We introduce a cyclic sampling strategy to solve this problem. All keyframe indexes are incremented by one in the next timestep. Taking six frames with two keyframes as a toy example, keyframe indexes at the first three timesteps are $(1, 4)$, $(2, 5)$, $(3, 6)$. Starting from the $4$-th timestep, the key frame selection will cycle from $(1, 4)$ again. In this way, keyframes and non-keyframe are evenly selected to prevent visual discrepancy, while we only have three inter-frame pairs that need to 
store inter-frame parameters.

\textbf{Keyframe manipulation.}~Video frames are divided into multiple $N$-frame batches. For each batch, in every sampling timestep, 
we cyclically select keyframes to apply our \METHODNAME-guided video manipulation. As with Sec.~\ref{sec:ebsynth}, we introduce the first and the last frames in each batch as anchor frames.
To impose intra-batch consistency, the anchor frames are always selected as the keyframes. 
To impose inter-batch consistency, the anchor frames in the previous batch are always included in the keyframes of the current batch. 
The latent features of the anchor frames are stored to replace the corresponding latent features in the next batch.

\textbf{Non-keyframe interpolation.} 
TokenFlow propagates features after the self-attention layers, which is orthogonal to our \METHODNAME~adaptation. We directly apply TokenFlow to propagate features after our \METHODNAME-guided attention for non-keyframe interpolation.

\subsubsection{Remark}
\label{sec:three-level}
Our \METHODNAME~guidance can be flexibly integrated into image manipulation and frame interpolation pipelines.
This paper presents two standard combinations: 1) SDEdit + Ebsynth, 2) Plug-and-Play + TokenFlow. We also implement Plug-and-Play + Ebsynth and SDEdit + TokenFlow but do not notice obvious performance differences from the standard combinations. 
Clearly, \METHODNAME~is not limited to these. For example, we can build a three-level hybrid framework for very long video translation: apply \METHODNAME-guided SDEdit to the primary keyframes, apply \METHODNAME-adapted TokenFlow to interpolate the secondary keyframes, and further interpolate the remaining non-keyframes with Ebsynth. A 16-second video example is shown in the last row of Fig.~\ref{fig:teaser}.

\begin{figure*}[t]
\centering
\includegraphics[width=0.98\linewidth]{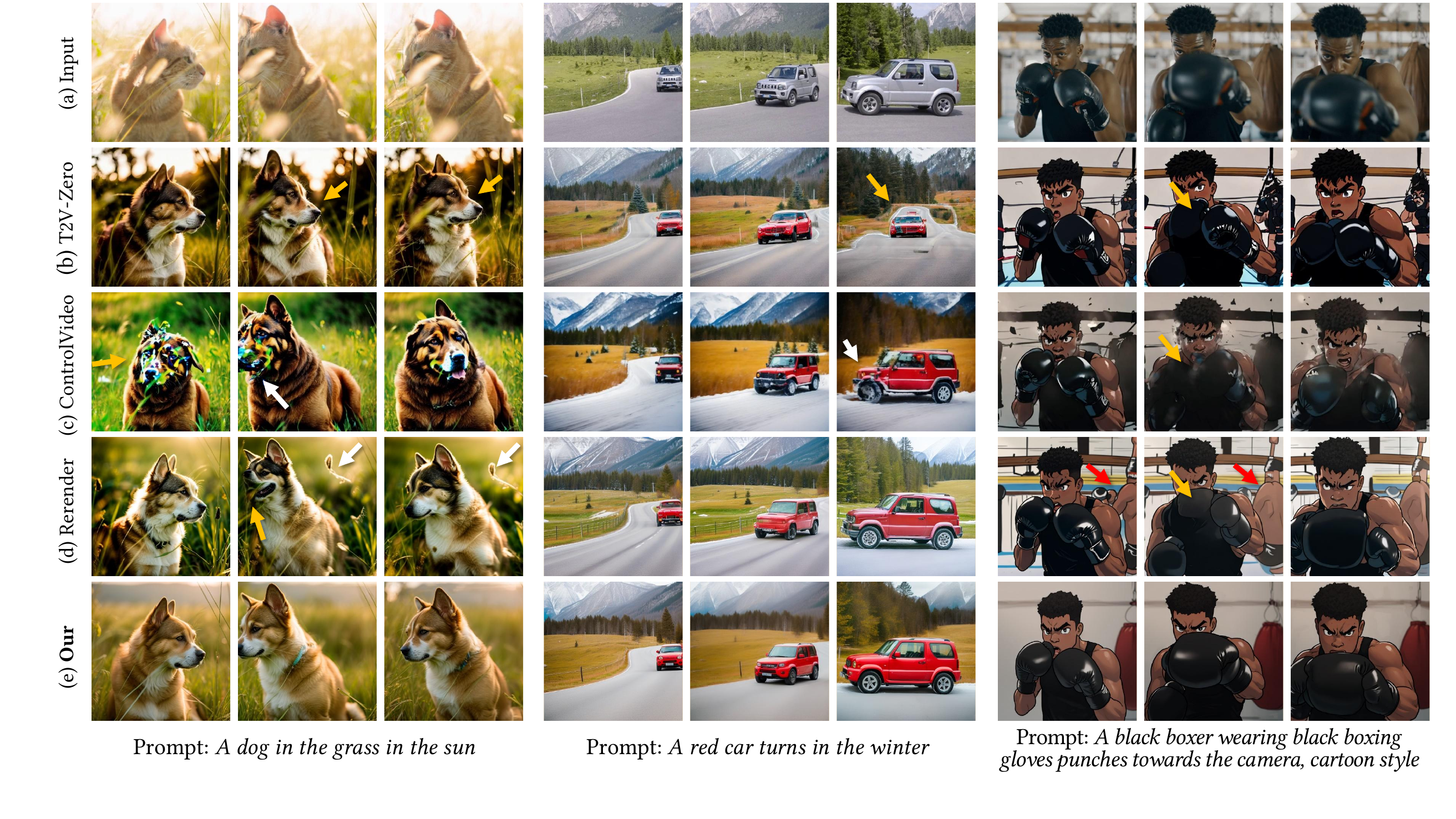}\vspace{-2mm}
\caption{Visual comparison with zero-shot video translation methods. Red, white, yellow arrows indicate the temporal inconsistency, artifacts, inconsistency with the input video, respectively.}\vspace{-3mm}
\label{fig:compare}
\end{figure*}

\begin{figure*}[t]
\centering
\includegraphics[width=0.98\linewidth]{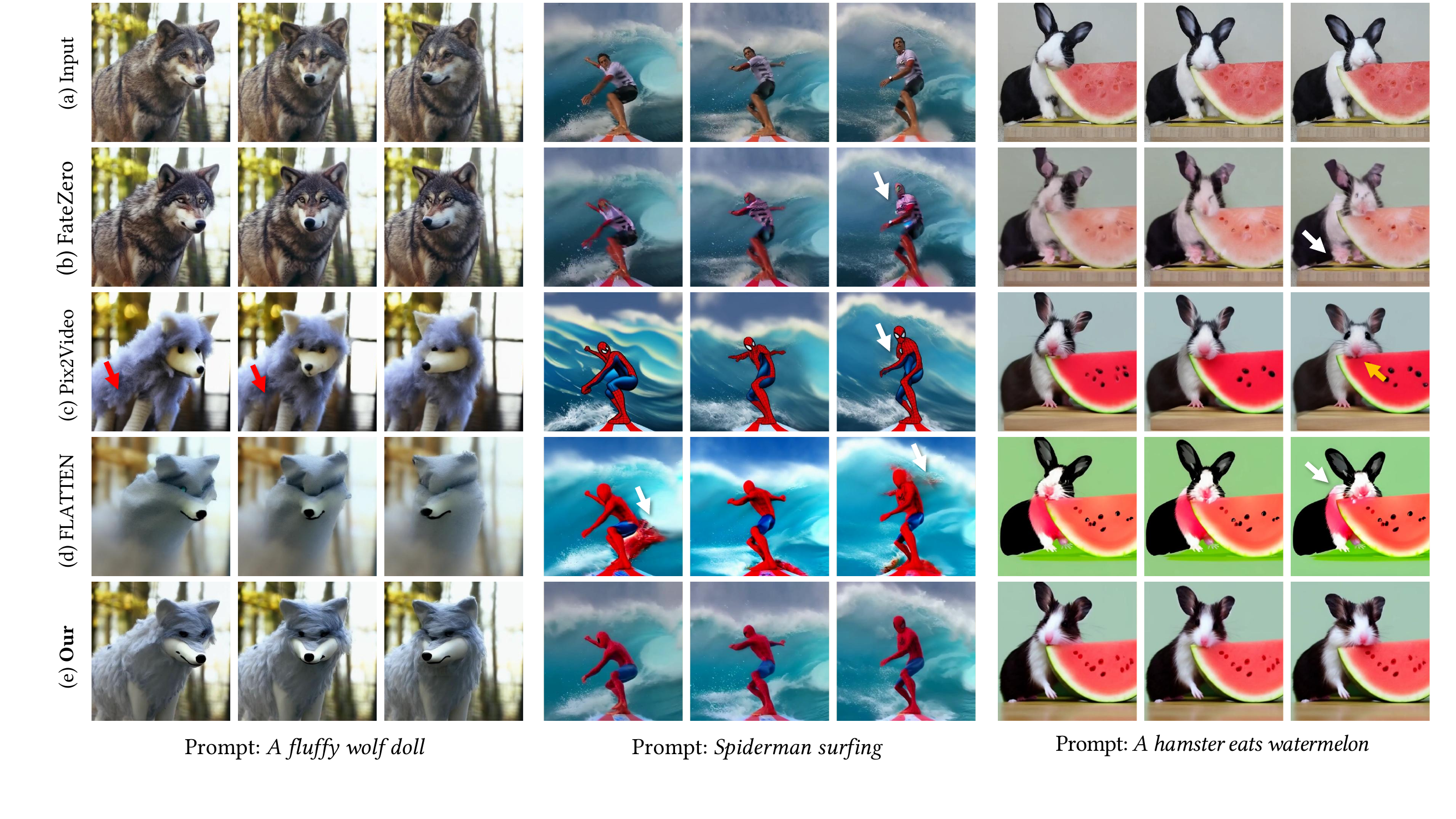}\vspace{-2mm}
\caption{Visual comparison with zero-shot text-guided video editing methods. Red and white arrows indicate the temporal inconsistency and  artifacts, respectively.}\vspace{-3mm}
\label{fig:compare2}
\end{figure*}

\section{Experiments}
\label{sec:experiment}

\subsection{Implementation Details}

We build our method on lightweight U-Net-based Stable
Diffusion models, which are capable of operating on a single 24-GB GPU.
By default, we set the batch size $N\in[6,8]$ based on the input video resolution. When combined with TokenFlow, we set batch size $16$ with $4$ keyframes. 
We set the loss weight $\lambda_\text{spat}=50$, the scale factors $\lambda_s=\lambda_t=5$. For feature optimization, we update $\mathbf{f}$ for $K=20$ iterations with Adam optimizer and a learning rate of $0.4$. We find that optimization mainly converges when $K=20$ and larger $K$ does not bring obvious gains.
GMFlow~\cite{xu2022gmflow} is used to estimate optical flows and occlusion masks.
Our \METHODNAME~adaptation takes less than additional 5 seconds and 1 GB memory per $512\times512$ frame. As a reference, the DDIM inversion process costs about 5 seconds per frame. Thus \METHODNAME~does not have excessive time and memory consumption.
Full details and results are provided in the supplementary material.

\subsection{Comparison with State-of-the-Art Methods}

\textbf{Zero-shot video-to-video translation.} 
We compare with three zero-shot video translation methods: Text2Video-Zero~\cite{khachatryan2023text2video}, ControlVideo~\cite{zhang2023controlvideo}, and Rerender-A-Video~\cite{yang2023rerender}.
To ensure a fair comparison, all methods employ identical settings of ControlNet, SDEdit, and LoRA.
In Fig.~\ref{fig:compare}, it is evident that all methods effectively translate videos based on the prompts. Nevertheless, the methods that rely on ControlNet conditions without inversion features can suffer a reduction in video editing quality, particularly if those conditions are degraded by factors such as defocus or motion blur. For instance, ControlVideo fails to render a plausible appearance of the dog or the boxer. Text2Video-Zero and Rerender-A-Video face challenges in preserving the cat's pose and the structural integrity of the boxer's gloves. In contrast, our method utilizes the robust guidance of \METHODNAME~to produce consistently reliable videos.

For quantitative evaluation, adhering to standard practices~\cite{qi2023fatezero, ceylan2023pix2video, yang2023rerender}, we employ the evaluation metrics of Fram-Acc (CLIP-based frame-wise editing accuracy),
Tmp-Con (CLIP-based cosine similarity between consecutive frames) and Pixel-MSE (averaged mean-squared pixel error between aligned
consecutive frames).
We also report Spat-Con ($L_1$ distance of the gram matrix on VGG features~\cite{gatys2016image}) for spatial coherence.
The results averaged over 23 videos are presented in Table~\ref{tb:quantitative_evaluation}. Importantly, our method achieves superior editing accuracy and temporal consistency. Additionally, a user study is conducted with 57 participants who are asked to choose the most preferred results from the four methods. Table~\ref{tb:quantitative_evaluation} displays the average preference rates across 11 test videos, showing that our method is the most favored.

\textbf{Zero-shot text-guided video editing.} 
We compare with three zero-shot video editing methods: FateZero~\cite{qi2023fatezero}, Pix2Video~\cite{ceylan2023pix2video}, FLATTEN~\cite{cong2023flatten} in Fig.~\ref{fig:compare2}.
FateZero fails to edit the target content while other methods successfully edit videos according to the text prompts provided.
Compared to inversion-free methods, inversion-based methods are good at preserving input structures
but may suffer artifacts when merging the inversion features and edited features. 
Pix2Video suffers from temporal inconsistency.
FLATTEN and our method achieve good temporal consistency and overall editing quality, and our results have fewer artifacts.

The results of the quantitative evaluation conducted on 11 test videos and 21 users in Table~\ref{tb:quantitative_evaluation} are consistent with the qualitative findings: FateZero essentially replicates the input videos, thus having the highest Spat-Con. Apart from this metric, our method achieves top scores in all other metric.

\begin{table} [t]
\caption{Quantitative evaluations on two tasks.}
\label{tb:quantitative_evaluation}
\resizebox{\linewidth}{!}{
\centering
\begin{tabular}{l|c|c|c|c|c}
\toprule
\multicolumn{6}{c}{\textbf{Video-to-Video Translation}}\\
\midrule
Metric & Fram-Acc $\uparrow$ & Tem-Con $\uparrow$ & Pixel-MSE $\downarrow$ & Spat-Con $\downarrow$  & User $\uparrow$  \\
\midrule
T2V-Zero & 0.918 & 0.965 & 0.038 & 388.3 & 9.1\% \\
ControlVideo & 0.932 & 0.951 & 0.066 & 389.2 & 2.6\% \\
Rerender & 0.955 & 0.969 & 0.016 & 400.6 & 23.3\% \\
Ours & \textbf{0.978} & \textbf{0.975} & \textbf{0.012} & \textbf{369.7} & \textbf{65.0\%}\\
\midrule
\multicolumn{6}{c}{\textbf{Text-Guided Video Editing}}\\
\midrule
Metric & Fram-Acc $\uparrow$ & Tem-Con $\uparrow$ & Pixel-MSE $\downarrow$ & Spat-Con $\downarrow$  & User $\uparrow$  \\
\midrule
FateZero & 0.341 & 0.985 & 0.012 & \textbf{173.7} & 6.5\% \\
Pix2Video & \textbf{1.000} & \textbf{0.988} & 0.023 & 421.1 & 20.8\% \\
FLATTEN & 0.889 & \textbf{0.988} & 0.013 & 295.1 & 22.1\% \\
Ours & \textbf{1.000} & \textbf{0.988} & \textbf{0.010} & 285.9 & \textbf{50.6\%} \\
\bottomrule
\end{tabular}}
\end{table}

\subsection{Ablation Study}

\begin{figure}[t]
\centering
\includegraphics[width=\linewidth]{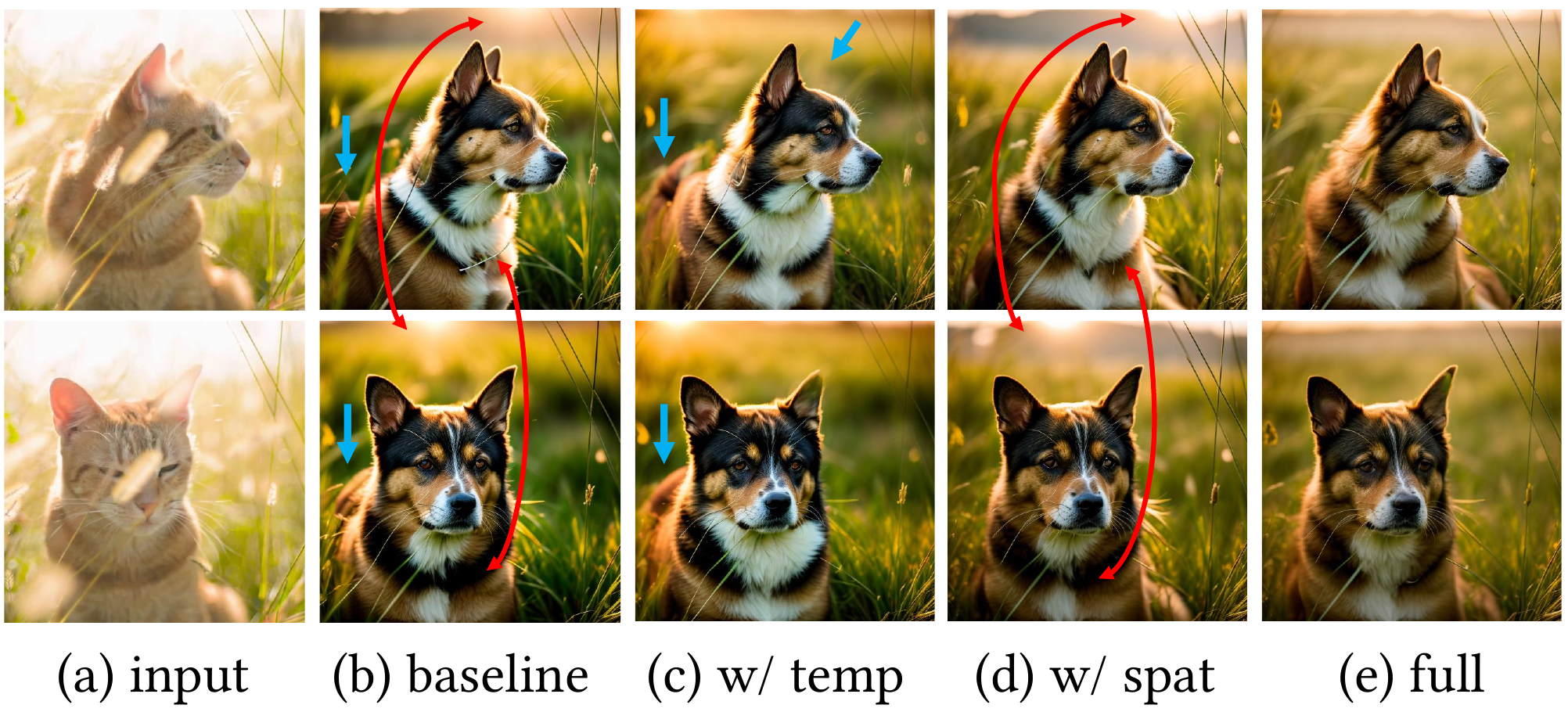}\vspace{-2mm}
\caption{Effect of incorporating spatial and temporal correspondences. The blue arrows indicate the spatial inconsistency with the input frames. The red arrows indicate the temporal inconsistency between two output frames.}
\label{fig:ablation1}
\end{figure}

To assess the impact of various modules on our framework's overall performance, we methodically deactivate certain modules. Figure~\ref{fig:ablation1} demonstrates the influence of integrating spatial and temporal correspondences. 
The baseline method relies solely on cross-frame attention to maintain temporal consistency. When temporal-related adaptation is applied, enhancements in consistency are evident, notably in texture alignment and sun's position between frames. Meanwhile, the spatial-related adaptation helps maintain the pose during translation.

\begin{table}[t]
\begin{center}
\caption{Quantitative ablation study on two tasks.}\vspace{-2mm}
\label{tb:quantitative_ablation}
\resizebox{\linewidth}{!}{
\centering
\begin{tabular}{l|cccccc}
\toprule
\multicolumn{7}{c}{\textbf{Video-to-Video Translation}}\\
\midrule
Metric & baseline & w/ temp & w/ spat & w/ attn & w/ opt & full\\
\midrule
Fram-Acc $\uparrow$ & \textbf{1.000} & \textbf{1.000} & \textbf{1.000} & \textbf{1.000} & \textbf{1.000} & \textbf{1.000} \\
Tem-Con $\uparrow$ & 0.974 & 0.979 & 0.976 & 0.976 & 0.977 & \textbf{0.980} \\
Pixel-MSE $\downarrow$ & 0.032 & 0.015 & 0.020 & 0.016 & 0.019 & \textbf{0.012} \\
Spat-Con $\downarrow$ & 528.0 & 433.1 & 404.1 & 410.3 & 425.4 & \textbf{355.6} \\
\midrule
\multicolumn{7}{c}{\textbf{Text-Guided Video Editing}}\\
\midrule
Metric & baseline & w/ temp & w/ spat & w/ attn & w/ opt & full\\
\midrule
Fram-Acc $\uparrow$ & 0.972 & \textbf{1.000} & 0.978 & 0.994 & 0.955 & \textbf{1.000} \\
Tem-Con $\uparrow$ & 0.976 & \textbf{0.989} & 0.984 & \textbf{0.989} & 0.988 & 0.988 \\
Pixel-MSE $\downarrow$ & 0.026 & 0.011 & 0.018 & 0.011 & 0.016 & \textbf{0.010} \\
Spat-Con $\downarrow$ & 319.4 & \textbf{284.2} & 300.8 & 290.9 & 303.1 & 285.9 \\
\bottomrule
\end{tabular}}\vspace{-2mm}
\end{center}
\end{table}

\begin{table} [t]
\caption{Effects of video interpolation with TokenFlow.}
\label{tb:tokenflow}
\resizebox{\linewidth}{!}{
\centering
\begin{tabular}{l|c|c|c|c}
\toprule
Metric & Fram-Acc $\uparrow$ & Tem-Con $\uparrow$ & Pixel-MSE $\downarrow$ & Spat-Con $\downarrow$   \\
\midrule
PnP+\METHODNAME & \textbf{1.000} & 0.988 & 0.010  & 285.9  \\
PnP+TokenFlow & 0.854  & \textbf{0.990}  & 0.007 & 252.0 \\
PnP+TokenFlow+\METHODNAME & 0.908 & \textbf{0.990} & \textbf{0.006} & \textbf{250.7}  \\
\bottomrule
\end{tabular}}
\end{table}

\begin{figure}[tbp]
\centering
\includegraphics[width=\linewidth]{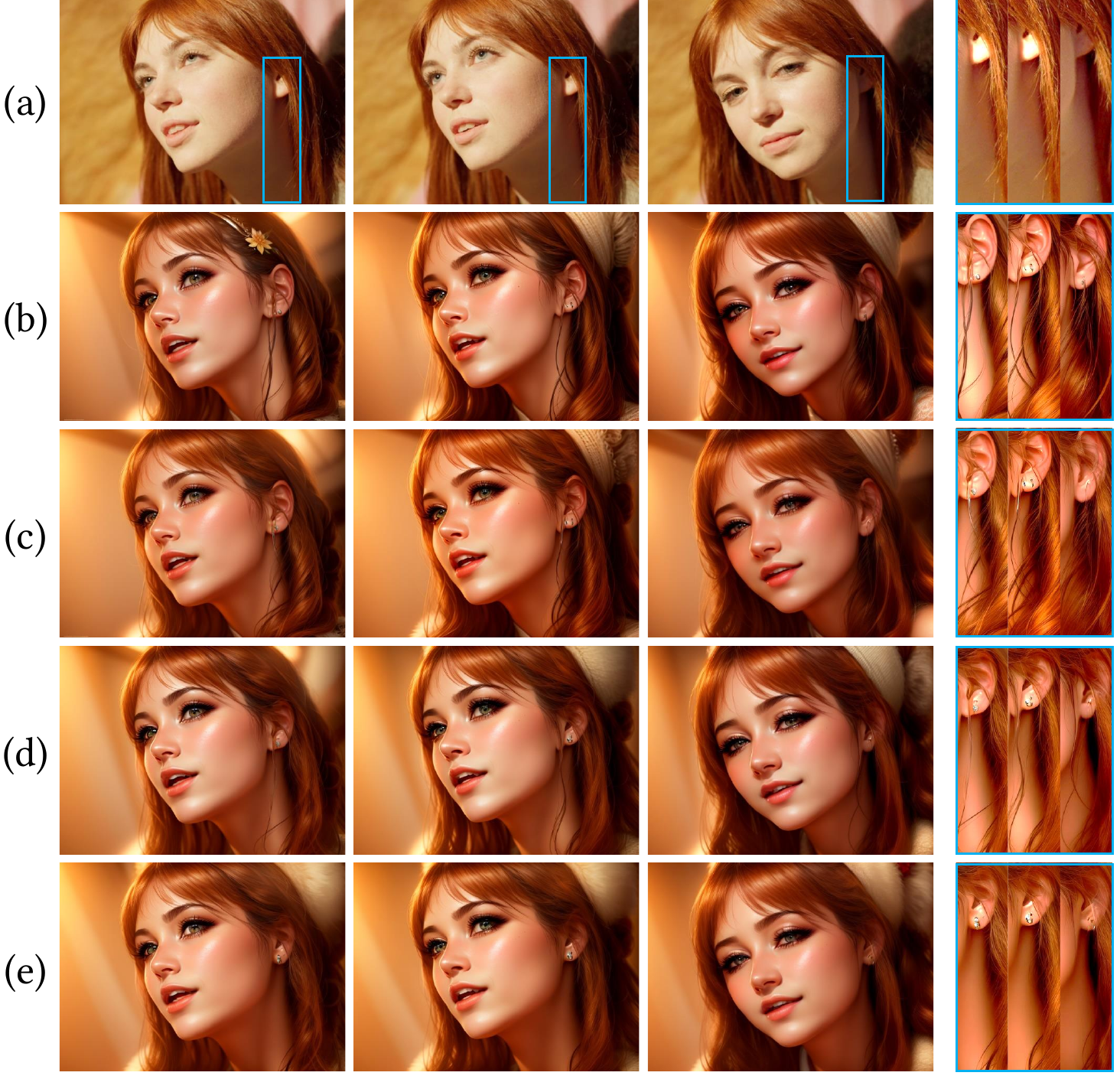}\vspace{-2mm}
\caption{Effect of attention adaptation and feature adaptation. Top row: (a) Input. Other rows: Results obtained with (b) only cross-frame attention, (c) attention adaptation, (d) feature adaptation, (e) both attention and feature adaptations, respectively. The blue region is enlarged with its contrast enhanced on the right for better comparison. Prompt: A beautiful woman in CG style.}
\label{fig:ablation_opt}
\end{figure}

\begin{figure}[t]
\centering
\includegraphics[width=\linewidth]{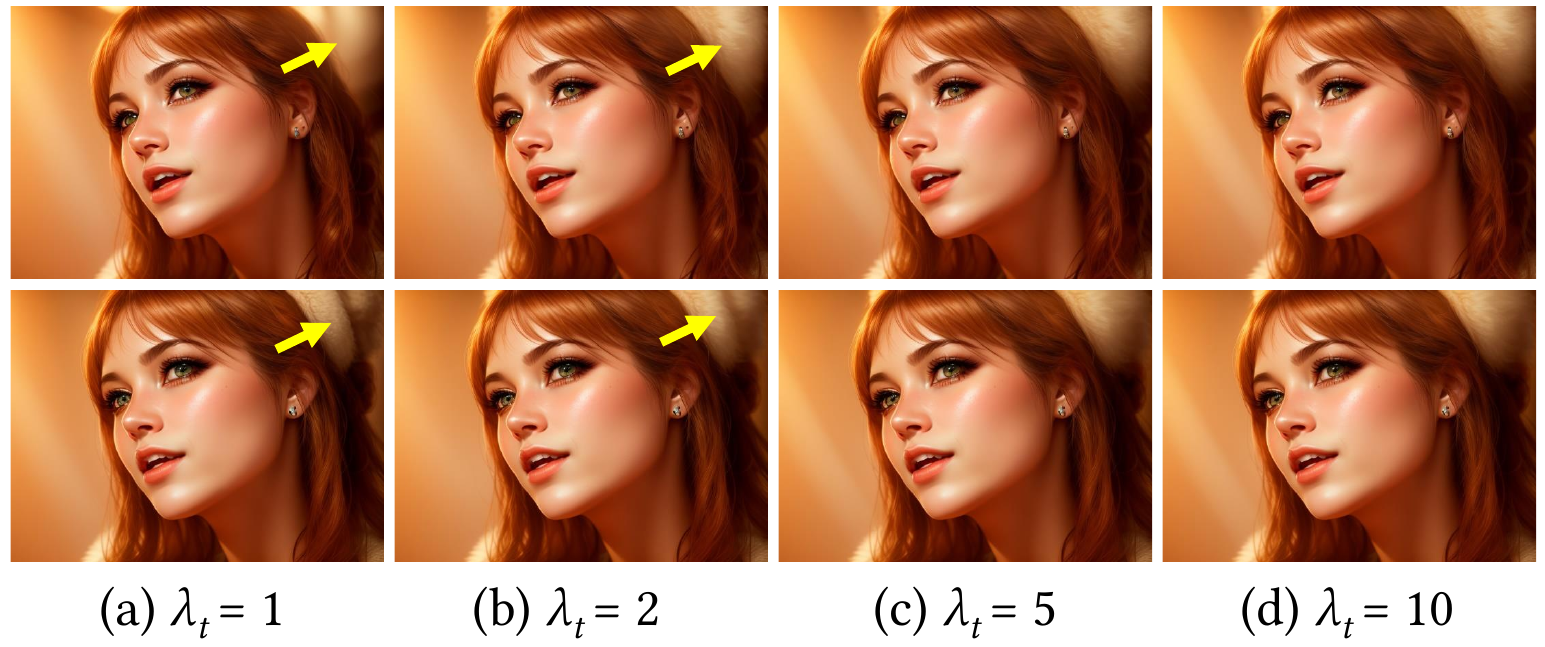}\vspace{-1mm}
\caption{Effect of $\lambda_t$. The yellow arrows indicate the inconsistency between the two frames.}\vspace{3mm}
\label{fig:ablation_lambdat}
\includegraphics[width=\linewidth]{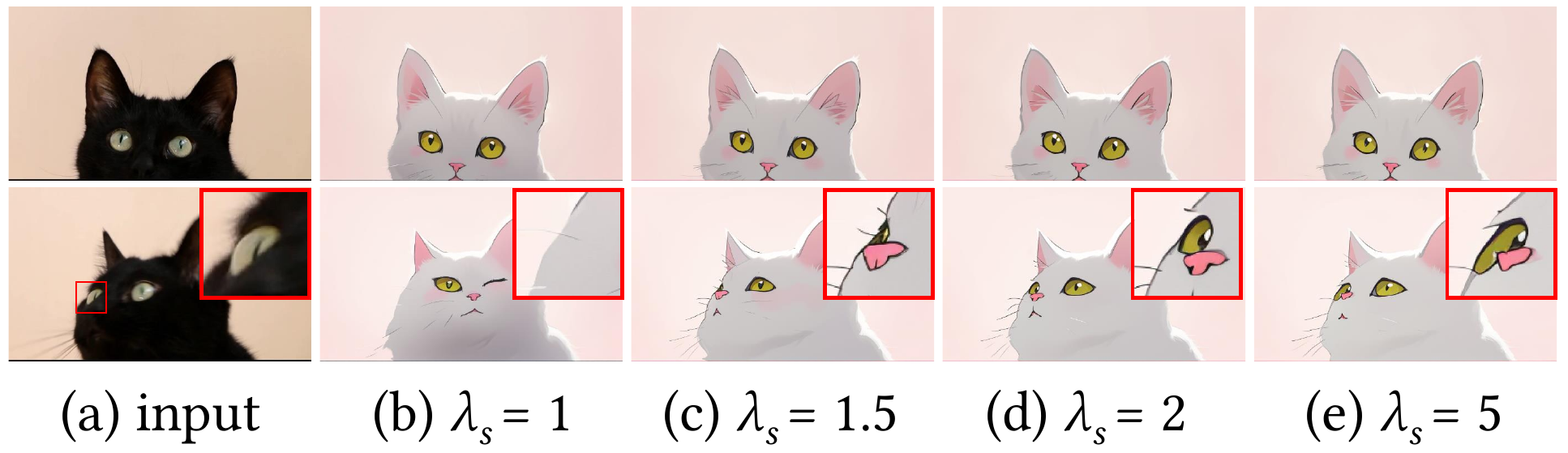}\vspace{-1mm}
\caption{Effect of $\lambda_s$. Red box region is enlarged and shown in the top right for better comparison. Prompt: A cartoon white cat in pink background.}\vspace{3mm}
\label{fig:ablation_lambdas}
\includegraphics[width=\linewidth]{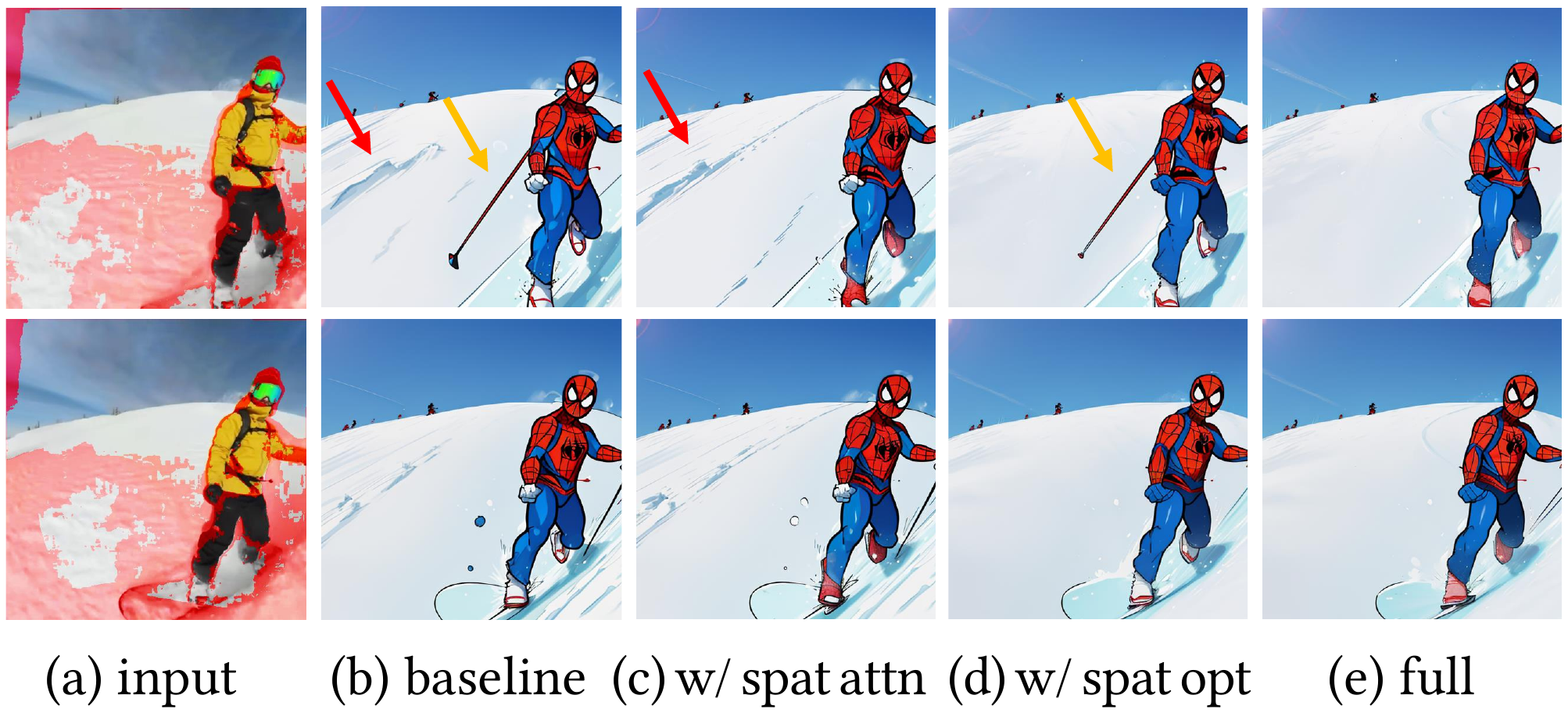}\vspace{-1mm}
\caption{Effect of spatial correspondence. (a) Input covered with red occlusion mask. (b)-(d) Our spatial-guided attention and spatial consistency loss help reduce the inconsistency in ski poles (yellow arrows) and snow textures (red arrows), respectively. Prompt: A cartoon Spiderman is skiing.}\vspace{3mm}
\label{fig:ablation_spat}
\includegraphics[width=\linewidth]{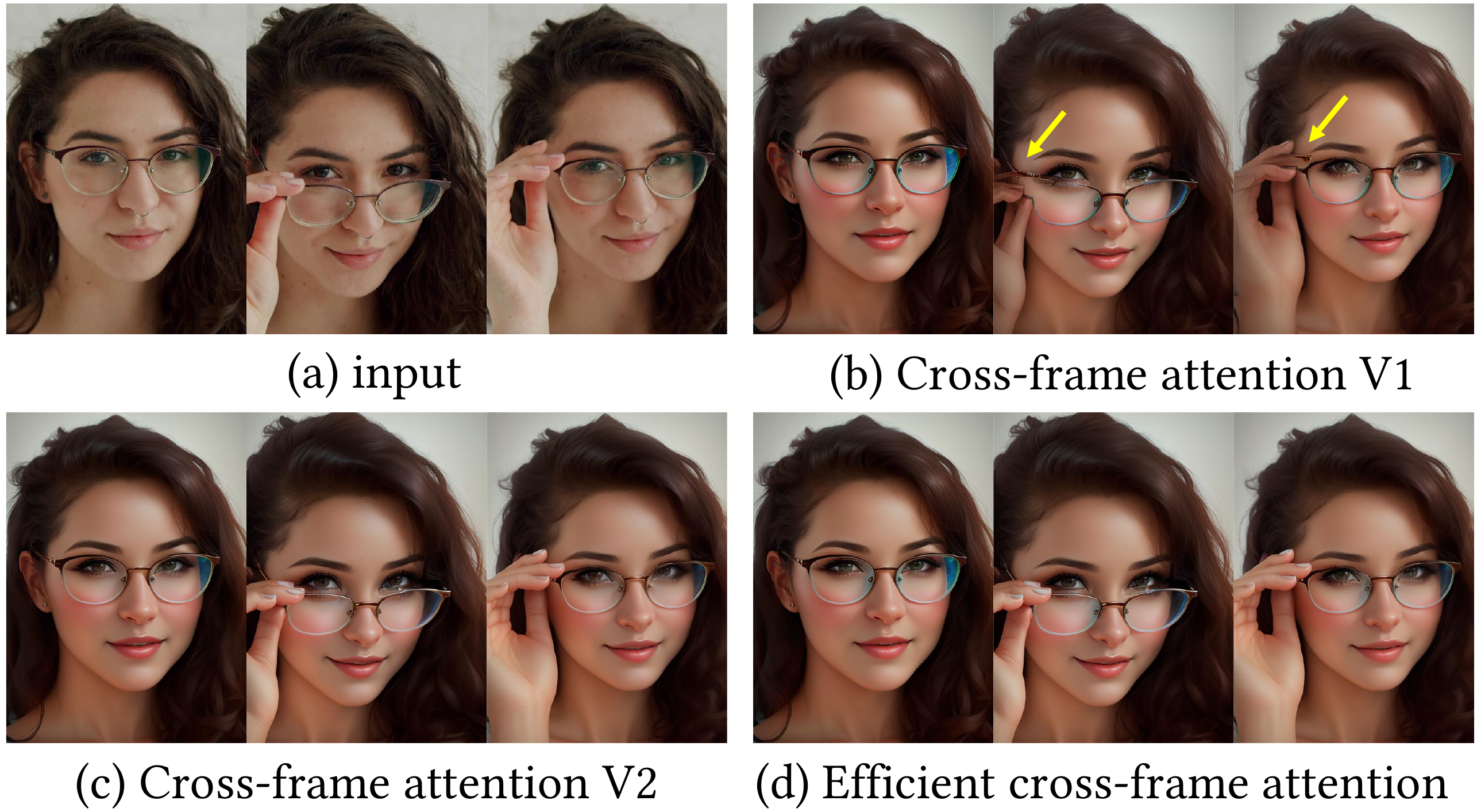}\vspace{-1mm}
\caption{Effect of efficient cross-frame attention. (a) Input. (b) Cross-frame attention V1 attends to the previous frame only, thus failing to synthesize the newly appearing fingers. (d) The efficient cross-frame attention achieves the same performance as (c) cross-frame attention V2, but reduces the region that needs to be attended to by $41.6\%$ in this example. 
Prompt: A beautiful woman holding her glasses in CG style.}
\label{fig:ablation_CF}
\end{figure}

Figure~\ref{fig:ablation_opt} examines the effect of attention and feature adaptation. Each modification alone could enhance temporal consistency, and combining both fully resolves the inconsistency in hair strands.
In terms of attention adaptation, we further study temporal- and spatial-guided attention. The strength of these constraints is governed by $\lambda_t$ and $\lambda_s$, respectively. As illustrated in Figs.~\ref{fig:ablation_lambdat}-\ref{fig:ablation_lambdas}, increasing $\lambda_t$ significantly improves background consistency, while increasing $\lambda_s$ enhances the alignment of the cat's pose between the transformed and original frames.
In addition to spatial-guided attention, our spatial consistency loss is crucial, as demonstrated in Fig.~\ref{fig:ablation_spat}. In this case, fast motion and blur lead to inaccurate optical flows and large occlusion regions. Spatial correspondence guidance is vital for guiding the rendering within this region. It is evident that each adaptation contributes, \eg, by removing an unwanted ski pole and fixing inconsistent snow textures. The combination of both provides the most coherent results.

Table~\ref{tb:quantitative_ablation} quantitatively evaluates the impact of each module.
In alignment with the visual results, it is evident that each module contributes to the overall enhancement of temporal consistency. Notably, the combination of all adaptations yields the best performance for video-to-video translation task.
For text-guided video editing task, since DDIM inversion already provides spatial guidance, introducing temporal guidance achieves best feature-level temporal consistency (Tem-Con) and Spat-Con. Nevertheless, our proposed spatial adaptation contributes to the best pixel-level temporal consistency (Pixel-MSE).
Table~\ref{tb:tokenflow} illustrates the effect of TokenFlow. It reveals that frame interpolation slightly lowers the editing success rate but greatly improves the temporal and spatial consistency. Our \METHODNAME~adaptation is highly compatible with TokenFlow to increase the editing success rate and further boost consistency.

\begin{figure}[tbp]
\centering
\includegraphics[width=\linewidth]{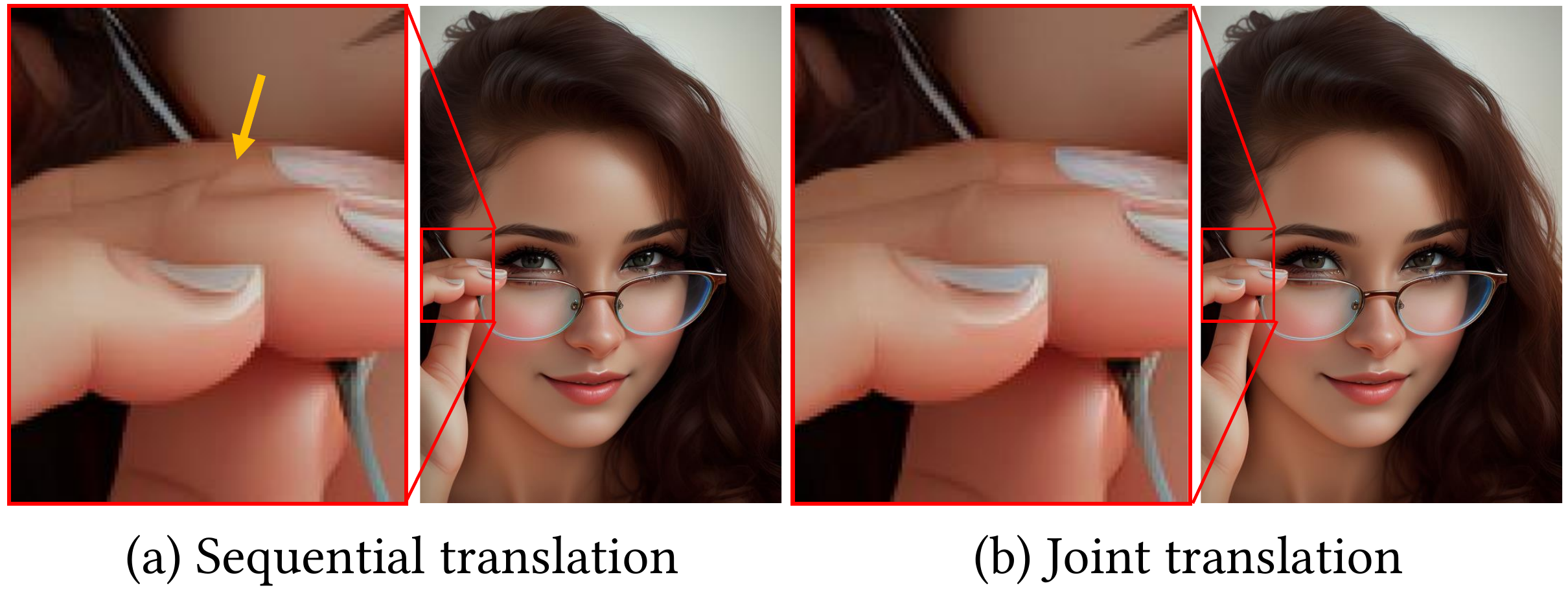}\vspace{-1mm}
\caption{Effect of joint multi-frame translation. Sequential translation relies on the previous frame alone. Joint translation uses all framesbioa batch to guide each other, thus achieving accurate finger structures by referencing to the third frame in Fig.~\ref{fig:ablation_CF}}\vspace{3mm}
\label{fig:ablation_joint}
\includegraphics[width=\linewidth]{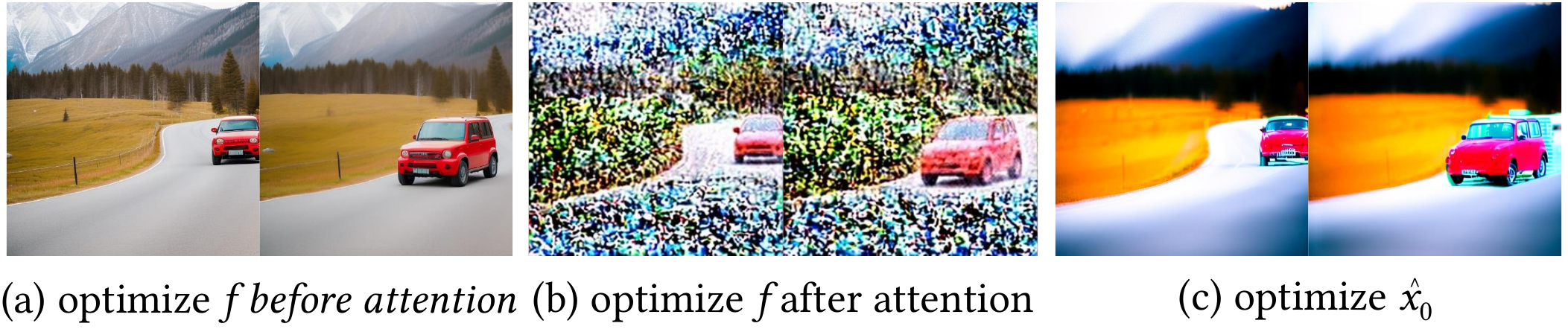}\vspace{-1mm}
\caption{Diffusion features to optimize.}\vspace{3mm}
\label{fig:ablation_optimize}
\includegraphics[width=\linewidth]{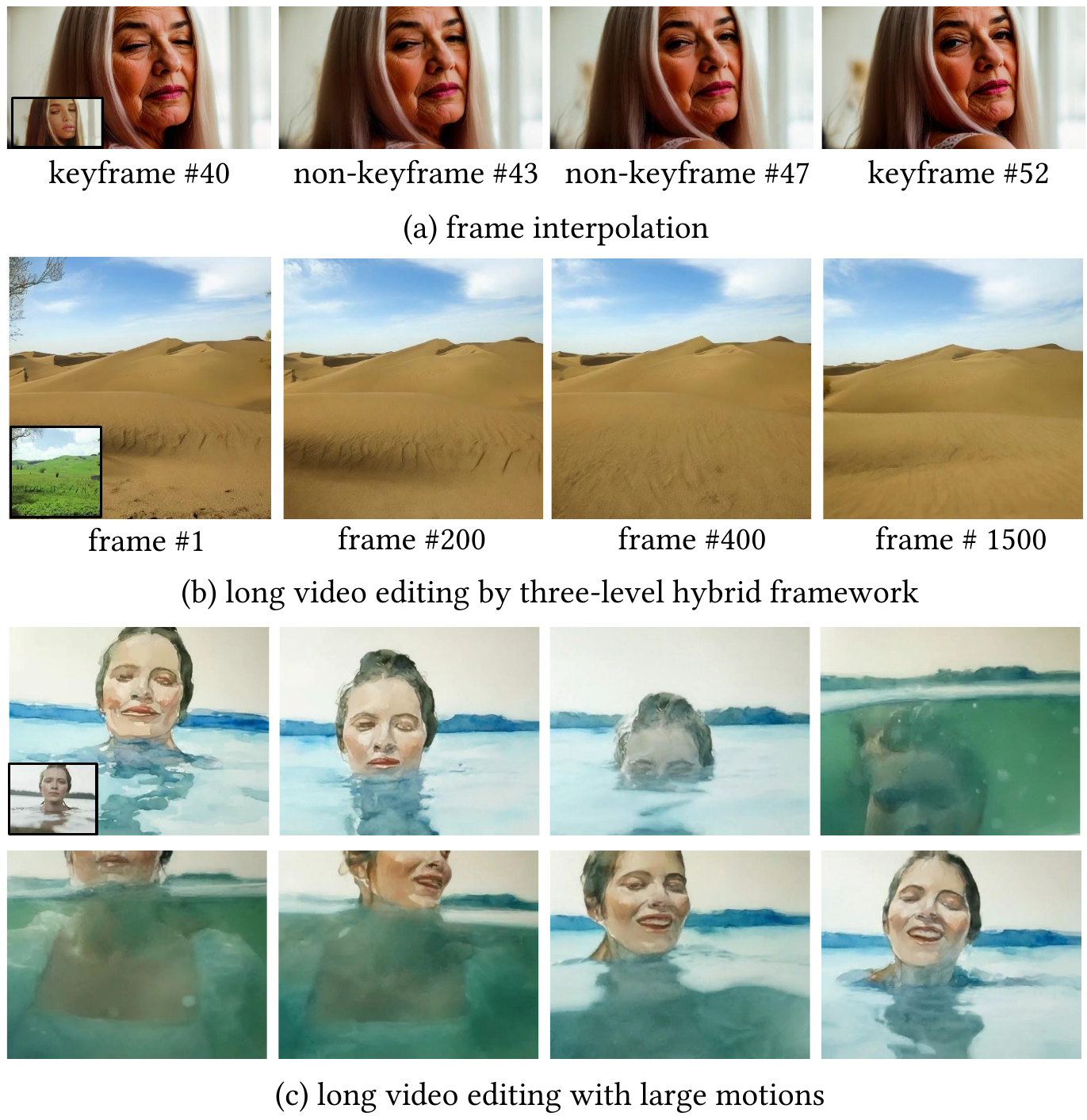}\vspace{-1mm}
\caption{Long video manipulation by our hybrid framework. (a) Prompt: A beautiful grandma, white hair. (b) Prompt: Desert, sand dunes and poplars. (c) A watercolor painting of a woman swimming in the water.}
\label{fig:long_video}
\end{figure}

Figure~\ref{fig:ablation_CF} ablates the proposed efficient cross-frame attention. As with Rerender-A-Video in Fig.~\ref{fig:challenge}(b), sequential frame-by-frame translation is vulnerable to new appearing objects. Our cross-frame attention allows attention to all unique objects within the batched frames, which is not only efficient but also more robust, as demonstrated in Fig.~\ref{fig:ablation_joint}.

\METHODNAME~uses diffusion features before attention layers for optimization. Since U-Net is trained to predict noise, the features after the attention layers (near the output layer) are noisy, leading to failure optimization (Fig.~\ref{fig:ablation_optimize}(b)). Meanwhile, the four-channel $\hat{x}'_0$ (Eq.~(\ref{eq:denoise})) is highly compact, which is not suitable for warping or interpolation. Optimizing $\hat{x}'_0$ results in severe blurs and over-saturation artifacts (Fig.~\ref{fig:ablation_optimize}(c)).

\subsection{Long Video Manipulation}

Figure~\ref{fig:teaser} presents examples of long video translation.~A 16-second video comprising 400 frames is processed, where 32 frames are selected as keyframes for diffusion-based translation and the remaining 368 nonkeyframes are interpolated. Thank to our \METHODNAME~guidance to generate coherent keyframes, non-keyframes exhibit coherent interpolation as in
Fig.~\ref{fig:long_video}(a).
Figure~\ref{fig:long_video}(b) further shows an example of a one-minute video with 1,500 frames, edited by our three-level hybrid framework as introduced in Sec.~\ref{sec:three-level}. When the camera moves forward all the time, our method can well preserve the shape of distant dunes. 
Figure~\ref{fig:long_video}(c) shows a 478-frame video where a woman quickly sinks into the water, then surfaces and turns her head. The scene changes rapidly and the motion is complex. Our method still works well on this challenging video, verifying the robustness of our method.
More video results are provided in the supplementary material.

\subsection{Limitation and Future Work}

In terms of limitations,
first, zero-shot methods rely on the editing ability of the image manipulation methods. Our method cannot handle the case where the image manipulation method fails to edit the target content. For example, we find that inversion-based image editing methods cannot turn the black cat white (\ie, Fig.~\ref{fig:ablation_lambdas}). The editing can only be realized by using the inversion-free ControlNet-based backbones.
Second, by enforcing spatial correspondence consistency with the input video, our method does not support large shape deformations and significant appearance changes.
The large deformation makes it challenging to use the optical flow of the original video as a reliable prior for natural motion. This limitation is inherent in zero-shot models. A potential future direction is to incorporate learned motion priors~\cite{guo2023animatediff}.
Third, this paper focuses on the lightweight U-Net-based Stable Diffusion models, which are capable of handling batched frames in a 24-GB consumer hardware GPU. Recent practices like Stable Diffusion 3~\cite{esser2024scaling} employ MM-DiT-based architecture for high-quality image generation. 
MM-DiT pachifies images into tokens and employs self-attention to handle two modalities, making it not directly compatible with our adaptation. Moreover, MM-DiT has a high hardware barrier compared to U-Net. 
It is challenging to run MM-DiT on a 24-GB consumer hardware GPU for multiple image generation. An interesting direction will be investigating the possibility of zero-shot video manipulation with MM-DiT on consumer hardware.

\section{Conclusion}\label{sec:conclusion}
This paper introduces a zero-shot framework for adapting image diffusion models to video manipulation. We illustrate the importance of maintaining intra-frame spatial correspondence within frames alongside inter-frame temporal correspondence, an aspect that has been underexplored in previous zero-shot methods. We further introduce a hybrid framework for long video manipulation. Our framework is instantiated for both inversion-free video-to-video translation and inversion-based text-guided video editing tasks. Our extensive experiments confirm the efficacy of our method in generating high-quality and coherent videos.

\bibliographystyle{IEEEtran}
\bibliography{bibliography}

\end{document}